%% file: Preprint2027.tex
\documentclass[letterpaper]{article}
\usepackage[preprint]{aaai2027}
\usepackage[hyphens]{url}
\usepackage{graphicx}
\usepackage{natbib}
\usepackage{caption}
\usepackage{amsmath}
\usepackage{amssymb}
\usepackage{algorithm}
\usepackage{algorithmic}
\usepackage{booktabs}
\newcommand{\best}[1]{{\bfseries\boldmath #1}}
\newcommand{\second}[1]{\underline{#1}}

\title{READER: Dynamic LLM Provenance from Query-Varying Interactions}
\author{
  Jiaxu Liu\textsuperscript{\rm 1,\rm 4}\equalcontrib,
  Sunnan Mu\textsuperscript{\rm 2}\equalcontrib,
  Dong Huang\textsuperscript{\rm 1},
  Liuyin Wang\textsuperscript{\rm 3},
  Jing Shao\textsuperscript{\rm 4},
  Jie Zhang\textsuperscript{\rm 4}\corresponding
}
\affiliations{
  \textsuperscript{\rm 1}National University of Singapore\\
  \textsuperscript{\rm 2}Xidian University\\
  \textsuperscript{\rm 3}Independent Researcher\\
  \textsuperscript{\rm 4}Shanghai Artificial Intelligence Laboratory\\
  \{jiaxu.liu,dong.huang\}@u.nus.edu,
  snmu@stu.xidian.edu.cn,
  liuyinwangthu@gmail.com\\
  \{shaojing,zhangjie1\}@pjlab.org.cn
}

\begin{document}

\maketitle

\begin{abstract}
Existing black-box LLM provenance methods achieve comparability by querying
every candidate model with the same diagnostic prompts. In deployment,
auditors inherit a different evidence stream: heterogeneous prompt--response
traces that arrive incrementally. We formalize \emph{dynamic black-box LLM
provenance}: after enrolling a fixed candidate ecosystem, attribute
query-varying interactions at any available evidence budget. READER recovers
comparability through a frozen proxy LLM. It projects proxy states aligned
with response tokens onto length-normalized DC and first-AC modes, capturing
response-wide activation location and coarse trajectory evolution. An
enrollment-trained linear probe converts each fingerprint into source
evidence, and Bayesian accumulation reuses this evidence unit from one
observation to many. We introduce Agent500, containing 50,000 responses from
100 local and API sources to 500 heterogeneous agent prompts. On 100-way
attribution, READER reaches $50.4\%$ accuracy from one response and $96.2\%$
from 100, compared with $33.0\%$ and $79.0\%$ for the strongest dynamic
baselines. Four distinct proxy families all exceed $94.8\%$ at the latter
budget. Controlled response-length and Math100 domain shifts expose the
limits of zero-retraining transfer. Component analysis reveals a
task-dependent spectral division of labor: DC dominates dynamic source
identity, while first AC dominates static relationship evidence. Their joint
fingerprint provides a shared measurement space for both tasks. READER audits
observed text without target internals or audit-only queries. Code and data
are available at
\url{https://github.com/LeoJeshua/READER}.
\end{abstract}

\input{main/introduction}
\input{main/related}
\input{main/method}
\input{main/experiments_before_analysis}
\input{main/experiments_after_analysis}
\input{main/analysis}
\input{main/conclusion}

\bibliography{aaai2027}

\clearpage
\appendix
\setcounter{secnumdepth}{3}
% AAAI maps \subsubsection to the subparagraph counter.
\renewcommand{\thesubparagraph}{\thesubsection.\arabic{subparagraph}}

\input{appendix/experimental_details}
\input{appendix/target_roster}
\input{appendix/algorithm}
\input{appendix/full_results}
\input{appendix/ablations_and_boundaries}
\input{appendix/ecosystem_geometry}
\input{appendix/cross_domain_geometry}
\input{appendix/reproducibility}

\end{document}

%% file: main/introduction.tex
\section{Introduction}
\label{sec:introduction}

Large language models increasingly operate as invisible infrastructure behind
agents, application backends, and third-party API services~\cite{
yang2026agentsurvey}. In this setting, the model name advertised by an endpoint
is not itself evidence of the model that served a request. An intermediary may
wrap a cheaper or unauthorized backend. A provider update may silently change
capability or safety behavior. A forensic investigation may need to associate
harmful output with its actual source. Model identity is therefore an
operational property rather than a documentation field, especially as API
behavior changes over time~\cite{tschisgale2026evidence}. An independent
auditor usually lacks parameters, logits, gradients, decoding traces, and
planted watermarks. The available evidence is the interaction itself: a prompt
and the text returned by an unknown backend.

\begin{figure*}[t]
  \centering
  \includegraphics[width=\textwidth]{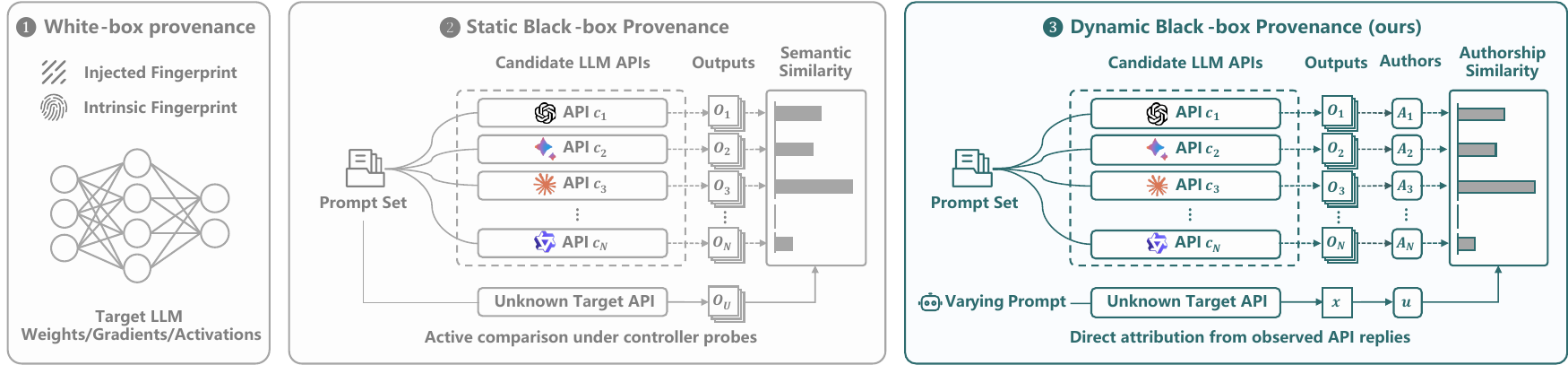}
  \caption{\textbf{Provenance settings.} White-box methods use target
  internals. Static black-box methods actively query candidate APIs with a
  shared panel. Dynamic provenance attributes observed, query-varying
  interactions after enrollment.}
  \label{fig:provenance_settings}
\end{figure*}

Intrinsic black-box methods demonstrate that model behavior can reveal
identity or lineage, but they collect evidence through shared, crafted, or
otherwise predefined prompt panels~\cite{gubri2024trap,pasquini2025llmmap,
Nikolic2025Model,yax2024phylolm,wu2026llmdna}. This controlled protocol is
effective for one-off comparison, yet continuous auditing presents a different
evidence stream. Audit-only calls add latency and API charges. Fingerprints
must be refreshed as services evolve. Meanwhile, heterogeneous traces from
ordinary users and agents remain unused. We call the complementary setting
\emph{dynamic black-box LLM provenance}: after enrolling a fixed candidate
ecosystem, attribute the source of observed, query-varying interactions under
the query budget currently available. Attribution proceeds directly from
traffic already produced in use. Figure~\ref{fig:provenance_settings}
distinguishes this setting from white-box access and static black-box
comparison.

Dynamic provenance removes the shared prompt that normally makes model outputs
comparable. Each response may concern a different task, and the available
record may contain one interaction or hundreds. The representation must
separate source identity from changing query semantics. The decision rule must
then accumulate weak measurements as the record grows. Our key hypothesis is
that a frozen proxy LLM supplies a common measurement space. When the proxy
reads an interaction, its response-token trajectory can reveal
source-dependent structure obscured in surface text and sentence embeddings.

\begin{figure*}[t]
  \centering
  \includegraphics[width=\textwidth]{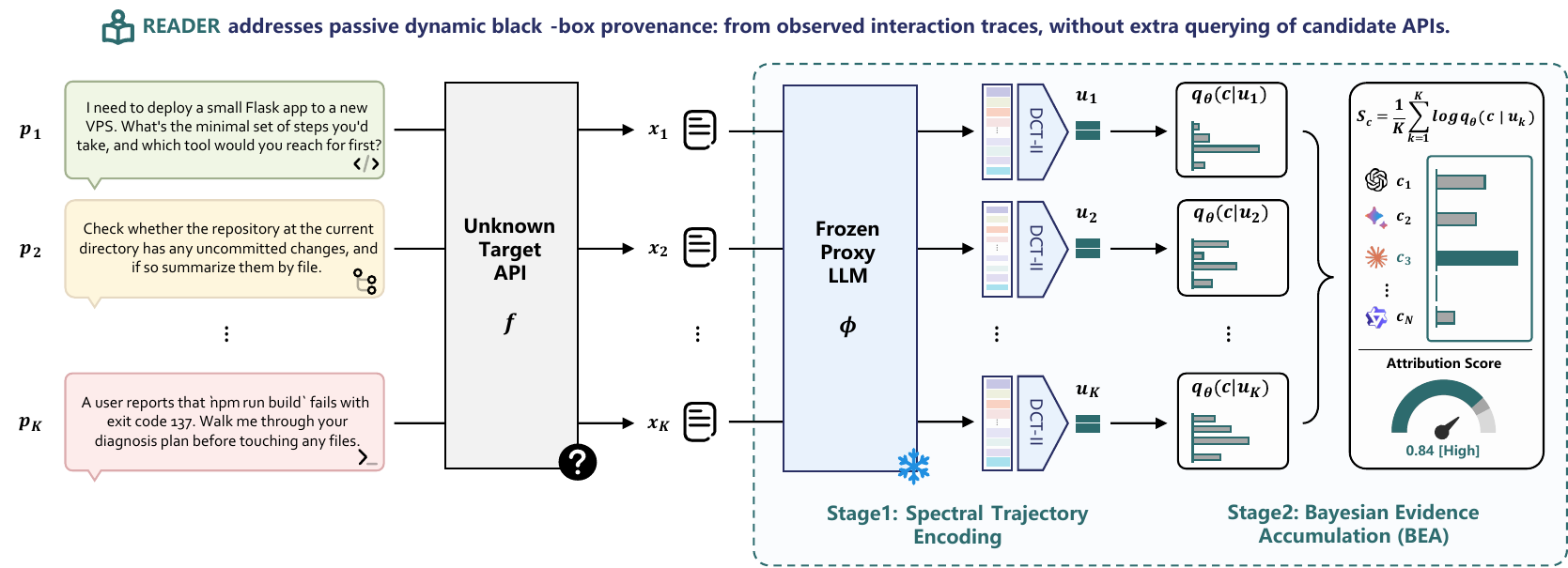}
  \caption{\textbf{READER pipeline.} A frozen proxy maps each interaction to
  a DC--AC trajectory fingerprint. One enrollment-trained probe reads
  response-level source evidence and accumulates it over any query budget.}
  \label{fig:reader_pipeline}
\end{figure*}

We instantiate this idea with \textbf{READER}, short for \emph{Response
Evidence for Authorship Decoding via Extracted Representations}
(Figure~\ref{fig:reader_pipeline}). READER retains the frozen proxy states
aligned with response tokens and applies a length-normalized DCT. The DC
coordinate captures response-wide activation location. The first AC coordinate
captures its coarsest zero-mean evolution. Together they form a trajectory
fingerprint that preserves two complementary views of authorship. A linear
probe converts one fingerprint into source evidence. Bayesian Evidence
Accumulation then combines log posteriors across observations, allowing one
enrollment-trained probe to serve every inference budget.

We evaluate this setting with \textbf{Agent500}, which crosses 500
heterogeneous agent prompts with 100 candidate sources, split evenly between
local and API models, to produce 50,000 interactions. READER reaches $50.4\%$
100-way accuracy from one response and $96.2\%$ from 100. The strongest
dynamic baselines reach $33.0\%$ and $79.0\%$, respectively. All four main
proxy families exceed $94.8\%$ at $K=100$. Controlled shortening and the
10,000 interaction \textbf{Math100} shift reveal where the enrolled evidence
stops transferring. Static relationship and original-space neighborhood tests
then show that the spectral coordinates preserve structure beyond a single
classification head. We will release both benchmarks, target metadata, and
the implementation.

Our contributions follow this progression. First, we formalize dynamic
black-box provenance from query-varying interactions and changing evidence
budgets. Second, we introduce a spectral proxy reader and a reusable
response-level evidence accumulator. Third, Agent500, Math100, and the
accompanying analyses establish 100-way attribution performance, expose its
transfer boundaries, and reveal model-ecosystem structure in the learned
measurement space.

%% file: main/related.tex
\section{Related Work}
\label{sec:related}

\paragraph{Provenance by design and internal access.}
Model identity can be established by planting a signal before deployment or by
comparing model internals after release. Decoding-time watermarks impose
detectable token-level statistics~\cite{kirchenbauer2023watermark,
kuditipudi2024robust}. Training-time backdoors in embedding services and
instruction-time fingerprints in generative LLMs create triggerable ownership
signatures~\cite{peng2023you,xu2024instructional}. Subsequent methods improve
semantic conditioning, scalability, and robustness to model merging~\cite{
gloaguen2026semanticallyconditioned,nasery2025scalablefingerprinting,
yamabe2025mergeprint}. A complementary white-box line derives fingerprints
from parameters, gradients, or internal representations~\cite{zeng2024huref,
wu2025gradient}. REEF, for example, compares suspect and reference activations
using centered kernel alignment~\cite{kornblith2019similarity,zhang2025reef}.
These methods establish provenance through release-time intervention or
suspect-model access. Dynamic black-box auditing begins after deployment, when
the interaction record is the available evidence.

\paragraph{Black-box behavioral provenance.}
Intrinsic black-box methods infer identity or lineage from a model's observable
behavior. Generated-text style and API-exposed logit spaces can expose model
fingerprints through model outputs~\cite{
mcgovern2024fingerprints,yang2024fingerprint}. TRAP and LLMmap construct
targeted or carefully crafted probes that amplify behavioral differences
between candidate models~\cite{
gubri2024trap,pasquini2025llmmap}. Model Provenance Testing and Model
Provenance Set formulate statistical decisions over candidate parents or
sources~\cite{Nikolic2025Model,qiu2026mps}. PhyloLM and LLM DNA build
functional representations for model similarity and evolutionary
relationships~\cite{yax2024phylolm,wu2026llmdna}. This literature establishes
the value of controlled queries for revealing provenance structure. Their
shared panels manufacture comparability across model outputs. Dynamic
provenance must recover that comparability from heterogeneous prompts already
observed in use, then strengthen a decision as additional interactions arrive.
READER addresses both requirements through a shared response fingerprint and
a budget-independent evidence unit.

\paragraph{Proxy representations as evidence.}
Linear probing has long shown that frozen neural states can make latent
attributes decodable. Probe controls clarify the distinction between
decodability and causal use~\cite{hewitt2019designing,ravichander2020probing}.
The linear
representation hypothesis and subsequent analyses characterize when such
structure can emerge in language-model representations~\cite{park2024linear,
jiang2024origins}. Superposition, dictionary learning, representation
engineering, and activation steering further show how behaviorally meaningful
attributes occupy high-dimensional activation space~\cite{
elhage2022toy,bricken2023monosemanticity,zou2023representation,
rimsky2024steering}. Concrete studies recover attributes such as linguistic
style and emotion~\cite{lai2024style,tak2025emotion}, with broader evidence
reviewed by \citet{zhang2026locate}. READER applies this representation-level
view across models. A frozen proxy contextualizes an observed interaction and
turns its response-token trajectory into an external measurement of source
behavior. Spectral coordinates retain response-wide location and coarse
tokenwise evolution. Their response-level readout supplies the evidence unit
needed for dynamic attribution.

%% file: main/method.tex
\section{Method}
\label{sec:method}

READER attributes black-box generations by using a frozen proxy language
model to read source-dependent evidence. Its two stages address the two
uncertainties of dynamic provenance. \emph{Spectral Trajectory Encoding}
extracts a comparable fingerprint from responses to different queries.
\emph{Bayesian Evidence Accumulation} converts each fingerprint into source
evidence and combines successive measurements under a changing query budget.

\subsection{Dynamic Provenance Setting}
\label{subsec:problem}

Let $\mathcal{C}=\{1,\ldots,C\}$ be a fixed ecosystem of candidate source
models. An unknown source $y\in\mathcal{C}$ receives prompts
$p_k\sim\mathcal{P}$ and generates responses
$x_k\sim P_y(\,\cdot\mid p_k)$, where $P_y$ is its conditional generation
distribution. Prompts may vary freely in task, domain, and semantic content.
Given a query budget $K$, dynamic provenance seeks
\begin{equation}
\widehat{y}
=
\mathop{\arg\max}_{c\in\mathcal{C}}
p\!\left(c\mid\mathcal{X}_{1:K}\right),
\qquad
\mathcal{X}_{1:K}=\{(p_k,x_k)\}_{k=1}^{K}.
\label{eq:dynamic_objective}
\end{equation}
Enrollment provides labeled responses from the candidate ecosystem.
Evaluation uses responses to unseen prompts. READER receives the observed
prompt and generated text as its complete audit record.

For each observed pair $(p,x)$, a frozen proxy model $\phi$ processes the
complete prompt and response. At proxy layer $\ell$, READER retains the hidden
states aligned with the response tokens:
\begin{equation}
\mathbf{H}^{(\ell)}(p,x)
=
\left[\mathbf{h}^{(\ell)}_0,\ldots,
\mathbf{h}^{(\ell)}_{T-1}\right]^\top
\in\mathbb{R}^{T\times d},
\label{eq:proxy_trajectory}
\end{equation}
where $T$ is the response length under the proxy tokenizer and $d$ is the
hidden dimension. The prompt supplies context to the proxy forward pass, but
only response-token states enter the trajectory encoder.

\subsection{Stage 1: Spectral Trajectory Encoding}
\label{subsec:spectral_encoding}

Query-varying prompts induce substantial semantic variation in responses, while
source-dependent evidence can persist in both the global location of a proxy
trajectory and its coarse evolution across tokens. READER isolates these
complementary signals by projecting all valid response states onto the first
two modes of a length-normalized DCT-II. For
$r\in\{0,1\}$, it computes
\begin{equation}
\mathbf{z}_{r}(p,x)
=
\frac{\gamma_r}{T}
\sum_{t=0}^{T-1}
\mathbf{h}^{(\ell)}_t
\cos\left[
\frac{\pi r}{T}\left(t+\frac{1}{2}\right)
\right],
\label{eq:length_normalized_dct}
\end{equation}
with $\gamma_0=1$ and $\gamma_1=\sqrt{2}$. Equation
\ref{eq:length_normalized_dct} is the orthonormal DCT-II coefficient divided
by $\sqrt{T}$, preventing response length from mechanically scaling its
magnitude.

The zeroth mode is the DC component,
\begin{equation}
\mathbf{z}_{0}(p,x)
=
\frac{1}{T}\sum_{t=0}^{T-1}\mathbf{h}^{(\ell)}_t,
\label{eq:dc_component}
\end{equation}
and records the response-wide activation location. The first AC component
$\mathbf{z}_1$ measures the smoothest front-to-back variation around that
location. Its temporal weights sum to zero, making it invariant to a constant
offset applied to the entire trajectory. For a single-token response,
$\mathbf{z}_1=\mathbf{0}$.
Modes $0$ and $1$ therefore form the minimal spectral pair that jointly retains
trajectory location and its lowest-order zero-mean evolution. Higher modes
encode progressively finer token-level variation and are studied separately
in the ablation.

% Terminology: use "DC--AC" in prose for the joint two-mode representation
% (an en dash, not arithmetic addition). Reserve "DC + first AC" for ablation
% labels only, and call mode r=1 the "first AC" rather than generic "AC".
The response fingerprint concatenates the two coefficients:
\begin{equation}
\mathbf{u}(p,x)
=
\begin{bmatrix}
\mathbf{z}_{0}(p,x)\\
\mathbf{z}_{1}(p,x)
\end{bmatrix}
\in\mathbb{R}^{2d}.
\label{eq:reader_fingerprint}
\end{equation}
This full-dimensional DC--AC representation is deterministic, uses every
response token, and leaves the proxy parameters fixed.
By preserving both absolute trajectory location and zero-mean evolution, the
same DC--AC fingerprint supports dynamic source attribution and static
relationship auditing through task-specific readouts. Stage 2 develops the
dynamic readout.

\subsection{Stage 2: Bayesian Evidence Accumulation}
\label{subsec:evidence_accumulation}

Each fingerprint is one noisy source measurement from a query-varying
response. READER calibrates these measurements into evidence over candidate
sources before combining them in decision space. Let
$\mathcal{D}_{\mathrm{tr}}=\{(\mathbf{u}_i,y_i)\}_{i=1}^{N}$ denote the
$N$ response-level examples in the enrollment training split, where
$\mathbf{u}_i=\mathbf{u}(p_i,x_i)\in\mathbb{R}^{2d}$ and
$y_i\in\mathcal{C}$. A coordinate-wise standardizer
$g_\psi:\mathbb{R}^{2d}\rightarrow\mathbb{R}^{2d}$ is fitted using only the
training split, with
$\psi=(\boldsymbol{\mu},\boldsymbol{\sigma})$ denoting its feature-wise means
and standard deviations and
$g_\psi(\mathbf{u})=(\mathbf{u}-\boldsymbol{\mu})
\oslash\boldsymbol{\sigma}$, where $\oslash$ denotes element-wise division.
On these standardized fingerprints, we fit a multinomial linear probe:
\begin{equation}
q_\theta(c\mid\mathbf{u})
=
\operatorname{softmax}
\left(\mathbf{W}^\top g_\psi(\mathbf{u})+\mathbf{b}\right)_c.
\label{eq:single_response_posterior}
\end{equation}
Here, $\mathbf{W}\in\mathbb{R}^{2d\times C}$,
$\mathbf{b}\in\mathbb{R}^{C}$, and $\theta=(\mathbf{W},\mathbf{b})$.
Because $\mathbf{u}(p,x)$ is computed from response states already
contextualized by $p$, we use $q_\theta(c\mid\mathbf{u}(p,x))$ as a
discriminative approximation to the prompt-conditioned source posterior.
The probe minimizes the regularized cross-entropy
\begin{equation}
\mathcal{L}(\theta)
=
-\frac{1}{|\mathcal{D}_{\mathrm{tr}}|}
\sum_{(\mathbf{u}_i,y_i)\in\mathcal{D}_{\mathrm{tr}}}
\log q_\theta(y_i\mid\mathbf{u}_i)
+
\frac{\lambda}{2}\|\mathbf{W}\|_F^2.
\label{eq:probe_objective}
\end{equation}
where $\|\cdot\|_F$ is the Frobenius norm and $\lambda\geq 0$ controls weight
decay. Response-level training makes $q_\theta$ the reusable evidence unit for
every inference-time query budget.

For a query set $\mathcal{X}_{1:K}$, write
$p_{1:K}=(p_1,\ldots,p_K)$, $x_{1:K}=(x_1,\ldots,x_K)$,
$\mathbf{u}_k=\mathbf{u}(p_k,x_k)$, and
$\mathbf{u}_{1:K}=(\mathbf{u}_1,\ldots,\mathbf{u}_K)$. We assume a stateless
query protocol: conditioned on the source and the observed prompts, target
requests do not share conversational or latent generation state. Therefore,
\begin{equation}
\begin{aligned}
p(x_{1:K}\mid p_{1:K},c)
&=
\prod_{k=1}^{K}p(x_k\mid p_k,c),
\\[-1pt]
\Longrightarrow\quad
p(\mathbf{u}_{1:K}\mid p_{1:K},c)
&=
\prod_{k=1}^{K}p(\mathbf{u}_k\mid p_k,c).
\end{aligned}
\label{eq:stateless_factorization}
\end{equation}
Conditioning on the observed prompts permits a heterogeneous prompt
distribution as long as prompt selection is source independent. Let
$\pi_c>0$ denote both the class prior represented in enrollment and the source
prior used at inference. Our class-balanced setting gives $\pi_c=1/C$.
Substituting
Eq.~\ref{eq:single_response_posterior} for the prompt-conditioned source
posterior and applying the corresponding prior correction yields the composite
log-posterior score
\begin{equation}
\begin{aligned}
S_c(\mathcal{X}_{1:K})
&=
\frac{1}{K}\sum_{k=1}^{K}
\log\!\left(q_\theta(c\mid\mathbf{u}_k)+\varepsilon\right)
\\
&\quad-
\frac{K-1}{K}\log\pi_c,
\end{aligned}
\label{eq:bayesian_evidence}
\end{equation}
where $\varepsilon>0$ provides numerical stability. READER predicts
\begin{equation}
\widehat{y}
=
\mathop{\arg\max}_{c\in\mathcal{C}}
S_c(\mathcal{X}_{1:K}).
\label{eq:reader_prediction}
\end{equation}
For a class-balanced ecosystem, the prior term is constant across sources, so
Eq.~\ref{eq:bayesian_evidence} becomes mean log-posterior accumulation. At
$K=1$, this is the single-response probe. Larger budgets reuse the same probe
and add evidence in log space.

The proxy layer $\ell$ and regularization strength $\lambda$ are selected using
a prompt-disjoint validation partition of the enrollment data and fixed before
test evaluation. Test prompts never enter fitting or model selection. Within
each split, enrollment data fit the standardizer, linear probe, and class
priors. The spectral coefficients cost $O(Td)$ per response and produce a
fixed $2d$-dimensional fingerprint. The dense probe costs $O(Cd)$ per
response, followed by $O(KC)$ accumulation. The frozen proxy forward pass
dominates inference cost in practice.

%% file: main/experiments_before_analysis.tex
\section{Experiments}
\label{sec:experiments}

\subsection{Experimental Setup}
\label{subsec:experimental_setup}

\paragraph{Tasks.}
Agent500 is our primary dynamic attribution benchmark. It contains 500
heterogeneous agent-style prompts, each answered by the same 100 candidate
source models (50 locally hosted and 50 accessed by API). Responses are
generated under a common temperature of $0.7$ and top-$p$ of $0.95$. Every
method observes the prompt--response pair. Math100 and controlled response
truncation test how the fixed Agent500 system transfers across domain and
length shifts. From Bench-A, introduced with MPT~\cite{Nikolic2025Model}, we
construct a balanced relationship-classification task over 116 model pairs
from 67 unique models. Positive pairs connect a parent to a derived model.
Negative pairs draw models from different parent families. Each model has 600
prompt-aligned responses.

\paragraph{Proxies and baselines.}
The main READER results use four instruction-tuned proxies from distinct
families: Llama-3.1-8B-Instruct, Gemma-4-12B-it,
Ministral-3-8B-Instruct-2512, and Qwen3.5-9B. We select one layer per proxy on
prompt-disjoint enrollment data and freeze it thereafter. Dynamic baselines
include random guessing, a fine-tuned DeBERTa-v3-large text classifier, and
dynamic adaptations of LLMmap~\cite{pasquini2025llmmap} and LLM-DNA with three
frozen sentence encoders~\cite{wu2026llmdna}. Their published protocols use a
fixed shared prompt panel, so Agent500 evaluates adapted response-level
versions under query-varying prompts. Static baselines are MPT,
PhyloLM~\cite{yax2024phylolm}, and the original LLM-DNA
concat--projection--centroid pipeline.

\paragraph{Protocol.}
Agent500 uses five-fold prompt-grouped cross-validation: all 100 responses to
one prompt remain in the same fold. Within each fold, the standardizer and
probe are fitted only on enrollment prompts. Query groups are sampled with
three fixed seeds for each $K\in\{1,5,10,20,50,100\}$. We report top-1
accuracy and macro-F1. Macro-F1 averages one-vs-rest F1 over all 100 sources,
giving equal weight to rare and difficult sources. The Bench-A-derived pair
task uses the same 20 fixed stratified $4{:}1$ splits for every method and
reports accuracy, precision, recall, F1, and AUC. The proxy-capability
diagnostic uses the official-style MMLU-Pro~\cite{wang2024mmlu} 5-shot CoT
protocol with non-thinking decoding.
Full rosters, hyperparameters, standard deviations, and per-seed results are
in the supplementary material.

\subsection{Dynamic 100-Way Attribution}
\label{subsec:dynamic_results}

\begin{figure*}[t]
  \centering
  \includegraphics[width=\textwidth]{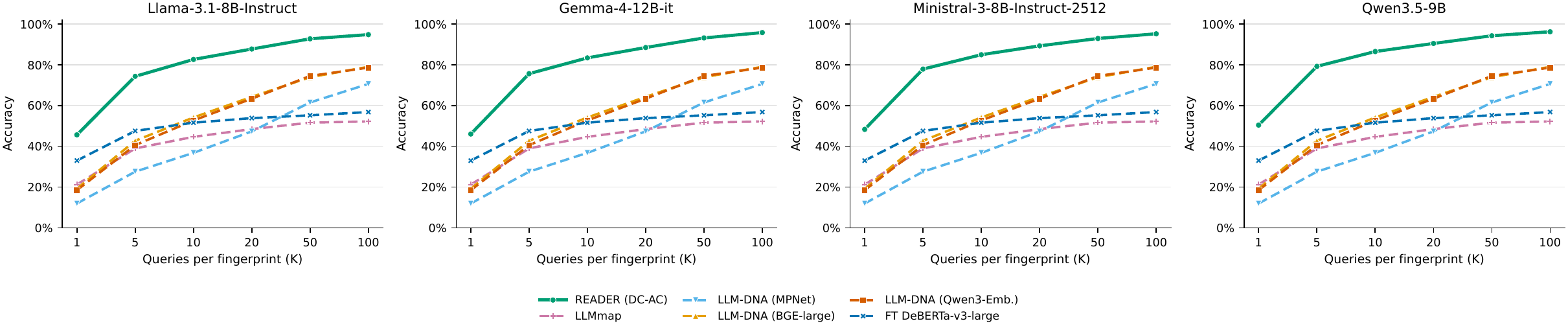}
  \caption{\textbf{Accuracy across query budgets on Agent500.} Each panel
  uses a distinct frozen proxy. A single response-level READER probe is reused
  for every $K$. Curves are means under the prompt-grouped protocol, with
  LLMmap and LLM-DNA adapted from their fixed-panel settings.}
  \label{fig:agent500_accuracy_vs_k}
\end{figure*}

Table~\ref{tab:agent500_main} gives the endpoint results for the two
requirements of dynamic provenance: reading evidence from one previously
unseen prompt and strengthening the decision as more heterogeneous queries
become available. Figure~\ref{fig:agent500_accuracy_vs_k} resolves the full
budget trajectory.

\begin{table}[!ht]
\centering
\small
\setlength{\tabcolsep}{2.4pt}
\begin{tabular*}{\columnwidth}{@{\extracolsep{\fill}}lcccc@{}}
\toprule
& \multicolumn{2}{c}{$K=1$} & \multicolumn{2}{c}{$K=100$} \\
\cmidrule(lr){2-3}\cmidrule(lr){4-5}
\textbf{Method} & \textbf{Acc.} & \textbf{M-F1} &
\textbf{Acc.} & \textbf{M-F1} \\
\midrule
Random                                & .010 & .010 & .010 & .007 \\
LLMmap\textsuperscript{\dag}          & .212 & .189 & .522 & .423 \\
DNA / MPNet\textsuperscript{\dag}     & .119 & .109 & .706 & .644 \\
DNA / BGE\textsuperscript{\dag}       & .191 & .188 & .790 & .739 \\
DNA / Qwen-Emb.\textsuperscript{\dag} & .185 & .192 & .786 & .735 \\
FT DeBERTa                            & .330 & .295 & .568 & .471 \\
\midrule
READER / Llama            & .456 & .455 & .948 & .931 \\
READER / Gemma            & .460 & .457 & \second{.958} & \second{.945} \\
READER / Ministral        & \second{.483} & \second{.481} & .952 & .939 \\
READER / Qwen             & \best{.504} & \best{.503} &
                            \best{.962} & \best{.952} \\
\bottomrule
\end{tabular*}
\caption{\textbf{Agent500 100-way attribution.} Means over five
prompt-grouped folds. M-F1 denotes macro-F1.
\textsuperscript{\dag}Dynamic adaptation of a fixed-panel method.}
\label{tab:agent500_main}
\end{table}
% Sources:
% outputs/exp_agent_n500/100-way/dct_alltokens_fullrank_best_layer_v2/reports/agent500/
% outputs/exp_agent_n500/100-way/llmmap_closed_q1_bea_v1/
% outputs/exp_agent_n500/100-way/reports_agg_logposterior/
% outputs/exp_agent_n500/100-way/finetuned_text_classifier/deberta_v3_large_generalization_v2/

From a single response, READER-Qwen reaches $50.4\%$ accuracy, a $17.4$-point
gain over the fine-tuned DeBERTa classifier and more than twice LLMmap. The
single-response result generalizes across readers, with all four READER
variants exceeding $45\%$. Accumulation then resolves sources that remain
ambiguous from one response. At $K=100$, READER-Qwen reaches $96.2\%$, which is
$17.2$ points above the strongest LLM-DNA adaptation. Every proxy exceeds
$94.8\%$. Figure~\ref{fig:agent500_accuracy_vs_k} traces the same advantage
through every intermediate budget. A probe trained once therefore converts a
growing interaction record into steadily stronger attribution.

\begin{figure}[!t]
  \centering
  \includegraphics[width=\columnwidth]{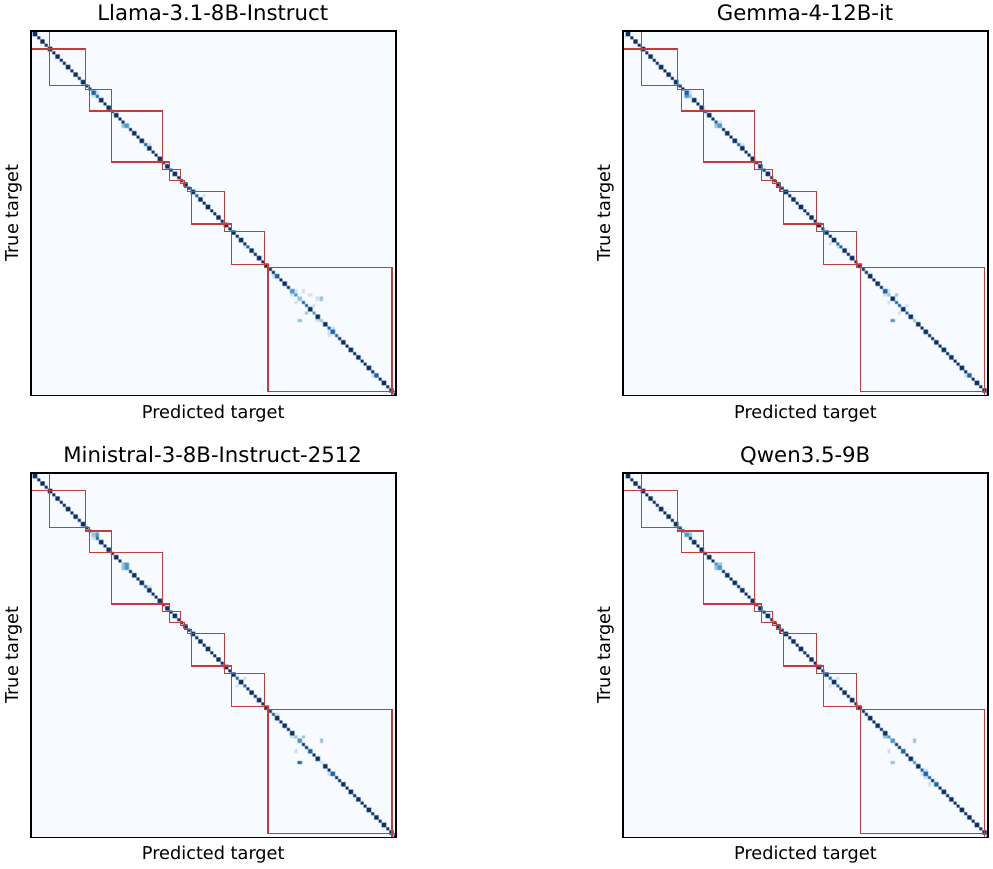}
  \caption{\textbf{Residual errors at $K=100$.} Row-normalized out-of-fold
  confusion matrices for the four main proxies. Targets are ordered by model
  family, and red boxes mark within-family blocks.}
  \label{fig:agent500_confusion}
\end{figure}
% Source:
% outputs/exp_agent_n500/100-way/dct_alltokens_fullrank_best_layer_v2/figures/ur_main/confusion_k100_main_2x2.pdf

\paragraph{Where do the remaining errors concentrate?}
Figure~\ref{fig:agent500_confusion} reveals a shared error geometry across
proxies. Llama, Gemma, Ministral, and Qwen make 26, 21, 24, and 19 errors,
respectively, among 500 out-of-fold decisions. Every error falls within a
family block, leaving no cross-family confusion. Accumulated evidence therefore
resolves family identity before separating closely related variants, which
remain the hardest attribution cases.

\begin{figure}[!ht]
  \centering
  \includegraphics[width=\columnwidth]{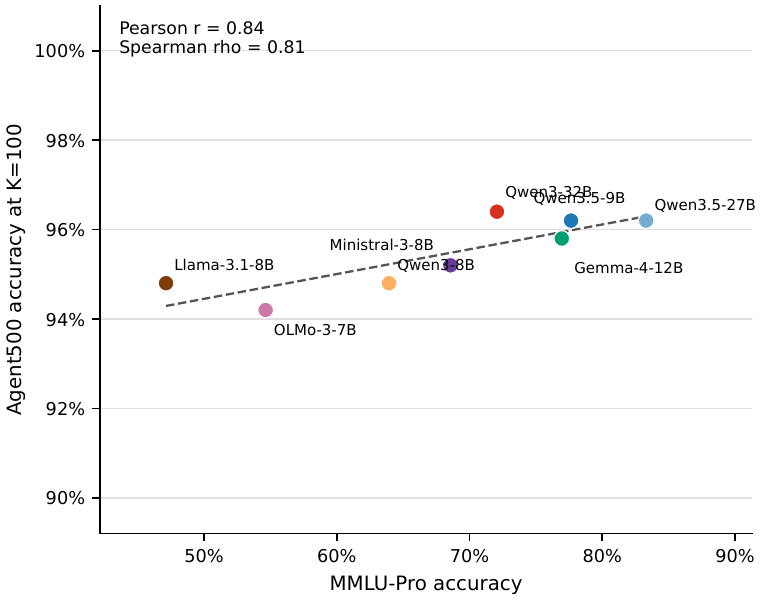}
  \caption{\textbf{Proxy capability and attribution.} MMLU-Pro accuracy
  versus Agent500 Acc.@100 across eight frozen proxies. The dashed line is a
  least-squares fit.}
  \label{fig:mmlu_proxy_correlation}
\end{figure}

\paragraph{Does a more capable proxy read stronger evidence?}
The consistency across proxy families raises a practical model-selection
question. Figure~\ref{fig:mmlu_proxy_correlation} compares MMLU-Pro accuracy
with Agent500 Acc.@100 over the full eight-proxy roster. The association is
strong in both value and rank. Pearson correlation is $r=0.84$ with $p=0.009$,
and Spearman correlation is $\rho=0.81$ with $p=0.015$. General language
competence is therefore a useful practical signal for proxy selection. We use
this eight-model association over near-ceiling Acc.@100 values as a diagnostic
of reader choice.

\subsection{Stress Tests and Method Boundaries}
\label{subsec:stress_tests}

We next freeze the Agent500-trained Qwen proxy and probe and change only the
evaluation responses. Table~\ref{tab:stress_tests} reports both absolute
performance and the fraction of clean Agent500 accuracy retained. The scores
measure zero-retraining transfer from the enrolled Agent500 distribution.

\begin{table}[!ht]
\centering
\small
\setlength{\tabcolsep}{4.0pt}
\begin{tabular*}{\columnwidth}{@{\extracolsep{\fill}}lccc@{}}
\toprule
\textbf{Evaluation condition} & \textbf{Acc.} & \textbf{M-F1} &
\textbf{Retained} \\
\midrule
Agent500 (clean)       & .962 & .952 & 100.0\% \\
Agent500 (64 tokens)   & .505 & .439 & 52.5\% \\
Agent500 (32 tokens)   & .159 & .111 & 16.5\% \\
Math100                & .330 & .242 & 34.3\% \\
\bottomrule
\end{tabular*}
\caption{\textbf{No-retraining stress tests at $K=100$.} A fixed
READER-Qwen probe trained on clean Agent500 is evaluated under response-length
and domain shifts. Retained is accuracy relative to clean Agent500.}
\label{tab:stress_tests}
\end{table}
% Sources:
% outputs/exp_agent_n500/100-way/dct_alltokens_fullrank_best_layer_v2/reports/length_ur/qwen35_9b.json
% outputs/exp_agent_n500/100-way/dct_alltokens_fullrank_best_layer_v2/reports/generalization_ur/qwen35_9b.json

READER retains useful but sharply reduced evidence under both shifts. At 64
tokens it achieves $50.5\%$ accuracy, compared with $27.8\%$ for fine-tuned
DeBERTa under the same truncation. At 32 tokens, their accuracy converges and
DeBERTa attains higher macro-F1. On Math100, READER reaches $33.0\%$ versus
DeBERTa's $23.2\%$, while retaining $34.3\%$ of its in-domain accuracy. The
pattern defines the transfer envelope of enrollment. Query content can vary
widely within Agent500, but severe shortening and domain shift remove much of
the learned evidence.

%% file: main/experiments_after_analysis.tex
\begin{figure*}[t]
  \centering
  \includegraphics[width=\textwidth]{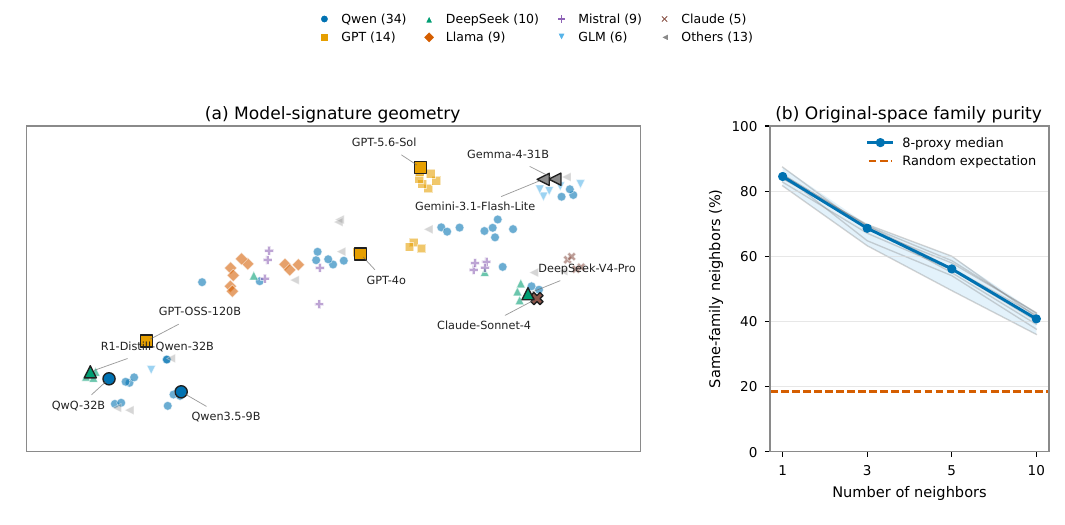}
  \caption{\textbf{Model-signature geometry.} (a) A sparse cosine t-SNE of
  100 Qwen-proxy signatures with representative stable neighborhoods labeled.
  (b) Same-family Top-$k$ purity measured before projection across eight
  proxies.}
  \label{fig:model_signature_geometry}
\end{figure*}

\subsection{Static Relationship Auditing}
\label{subsec:static_relationships}

Dynamic attribution establishes that the spectral coordinates separate
sources across unseen prompts. We next test whether they also preserve
relationships among sources on the balanced task derived from Bench-A. For
each pair, we compute prompt-aligned cosine similarity in the DC and first-AC
blocks, average both scores over prompts, and fit a linear SVM on the resulting
two-dimensional feature. This static readout is trained solely from the
derived pair labels.
Table~\ref{tab:bench_a_main} compares it with specialized relationship
baselines. Section~\ref{subsec:representation_analysis} then identifies which
spectral block supports each task.

\begin{table}[!ht]
\centering
\small
\setlength{\tabcolsep}{2.0pt}
\begin{tabular*}{\columnwidth}{@{\extracolsep{\fill}}lccccc@{}}
\toprule
\textbf{Method} & \textbf{Acc.} & \textbf{Prec.} & \textbf{Rec.} &
\textbf{F1} & \textbf{AUC} \\
\midrule
MPT                       & .742 & \second{.943} & .521 & .660 & .746 \\
PhyloLM                   & .758 & \best{.944} & .554 & .690 & .793 \\
DNA / MPNet               & .621 & .650 & .554 & .589 & .675 \\
DNA / BGE                 & .631 & .629 & \best{.646} & .631 & .722 \\
DNA / Qwen-Emb.           & .704 & .835 & .554 & .650 & .723 \\
\midrule
READER / Llama            & \best{.781} & .928 & .617 &
                            \best{.731} & \best{.828} \\
READER / Gemma            & .694 & .730 & \second{.629} & .668 & .768 \\
READER / Ministral        & .754 & .906 & .575 & .693 & \second{.811} \\
READER / Qwen             & \second{.767} & .907 & .600 &
                            \second{.714} & .809 \\
\bottomrule
\end{tabular*}
\caption{\textbf{Static relationship classification derived from Bench-A.}
READER uses the DC--AC two-score readout. Values are means over the same 20
stratified splits of 116 balanced model pairs.}
\label{tab:bench_a_main}
\end{table}
% Sources:
% outputs/bench_a_pairwise_blackbox_116/dct_component_grp128_comparison/
% outputs/bench_a_pairwise_blackbox_116/dct_alltokens_fullrank_best_layer_v2/reports/
% outputs/bench_a_pairwise_blackbox_116/reports/

\paragraph{Benchmark comparison.}
READER-Llama obtains the highest accuracy ($78.1\%$), F1 ($73.1\%$), and AUC
($82.8\%$). PhyloLM retains the highest precision. The result extends READER's
measurement space from source decisions to pairwise relationships, using the
same spectral coordinates with a task-specific linear readout.

%% file: main/analysis.tex
\subsection{Representation Analysis}
\label{subsec:representation_analysis}

Having established dynamic attribution and static relationship auditing, we
now ask what makes one representation serve both tasks.
Table~\ref{tab:representation_ablation} isolates its spectral coordinates and
the text presented to the proxy.

\begin{table}[!ht]
\centering
\small
\setlength{\tabcolsep}{1.5pt}
\begin{tabular*}{\columnwidth}{@{\extracolsep{\fill}}lcccc@{}}
\toprule
\multicolumn{5}{l}{\textit{(a) Spectral components across tasks}} \\
\addlinespace[2pt]
& \multicolumn{2}{c}{\textbf{Agent500}} &
  \multicolumn{2}{c}{\textbf{Pair task}} \\
\cmidrule(lr){2-3}\cmidrule(lr){4-5}
\textbf{Fingerprint} & \textbf{Acc.@1} & \textbf{Acc.@100} &
\textbf{Acc.} & \textbf{AUC} \\
\midrule
DC only               & .475 & \best{.956} & .541 & .621 \\
First AC only         & .383 & .939 & \best{.749} & .800 \\
DC--AC                & \best{.476} & .955 & \best{.749} & \best{.804} \\
\midrule
\multicolumn{5}{l}{\textit{(b) Proxy input on Agent500}} \\
\addlinespace[2pt]
\textbf{Proxy input} & \textbf{Acc.@1} & \textbf{F1@1} &
\textbf{Acc.@100} & \textbf{F1@100} \\
\midrule
Response only ($r$)          & \best{.481} & \best{.479} & .954 & .953 \\
Prompt--response ($ur$)      & .476 & .474 & \best{.955} & \best{.954} \\
\bottomrule
\end{tabular*}
\caption{\textbf{Representation analysis.} Four-proxy means. Panel (a)
compares spectral blocks across dynamic and static tasks. Panel (b) changes
the proxy input while encoding response-aligned states. F1 is macro-F1.}
\label{tab:representation_ablation}
\end{table}
% Sources:
% outputs/exp_agent_n500/100-way/dct_alltokens_fullrank_best_layer_v2/reports/agent500/
% outputs/bench_a_pairwise_blackbox_116/dct_component_grp128_comparison/summary.json

\paragraph{Complementary spectral coordinates.}
The roles reverse across tasks. DC dominates dynamic attribution and matches
the joint fingerprint within $0.1$ point at $K=100$. On the Bench-A-derived
pair task, first AC improves over DC by $20.8$ accuracy points. The joint
readout matches AC accuracy and yields the highest AUC. Global trajectory
location therefore carries most closed-set identity, whereas coarse zero-mean
evolution expresses pairwise relationships. Their union creates a shared
coordinate system in which lightweight readouts select the evidence relevant
to each task.

\paragraph{Proxy input.}
Response-only input is $0.5$ point better at $K=1$, whereas prompt--response is
$0.1$ point better at $K=100$. Macro-F1 follows the same near tie. Prompt
context therefore changes neither endpoint materially, locating the decodable
signal in response-token dynamics. The supplement decomposes source and prompt
variance under both inputs.

\subsection{Model-Signature Geometry}
\label{subsec:model_signature_geometry}

The static pair task reads relationships from labels. We finally remove that
readout and inspect the geometry directly. We average each source's 500
DC--AC fingerprints, normalize the two blocks, and compute cosine distance.
Figure~\ref{fig:model_signature_geometry} uses t-SNE for display and evaluates
neighborhoods in the original space.

Across eight proxies, median Top-1/Top-5 family purity is $84.5\%/56.1\%$
against an $18.4\%$ random expectation. Enrichment is significant for every
main proxy after Holm correction ($p\leq8.0\times10^{-4}$). The main proxies'
distance matrices also agree (median Spearman $0.818$).
DeepSeek-R1-Distill-Qwen-32B and QwQ-32B are mutual Top-10 neighbors under all
eight proxies, consistent with shared lineage. Cross-provider proximity
is interpreted as behavioral similarity within READER's measurement space.
The supplement reports full labels, prompt-subset and projection sweeps, and
all neighborhood checks.

%% file: main/conclusion.tex
\section{Limitations and Conclusion}
\label{sec:conclusion}

\paragraph{Limitations.}
READER assumes a fixed enrolled ecosystem, stateless requests, and
source-independent prompt selection. It cannot reject unknown sources or
model drift, and proxy inference adds cost. Table~\ref{tab:stress_tests} shows
substantial evidence loss under severe response shortening and domain shift.
Static auditing requires aligned prompts and a separate readout. Open-set,
drift-aware, and domain-robust attribution remain deployment challenges.

\paragraph{Conclusion.}
READER turns heterogeneous API interactions into cumulative provenance
evidence. A frozen proxy supplies a common measurement space, spectral
coordinates expose source structure, and one probe serves every query budget.
On Agent500, these fingerprints support dynamic attribution across 100
enrolled sources. Separate static analyses show that the same coordinates also
capture pairwise model relationships and ecosystem geometry. Once candidates
are enrolled, ordinary traffic becomes an audit record.

%% file: appendix/experimental_details.tex
\section{Data and Evaluation Protocol}
\label{app:experimental_details}

\subsection{Agent500 Construction}

Agent500 crosses 500 heterogeneous agent-style prompts with a fixed ecosystem
of 100 target models, yielding 50,000 prompt--response records. The ecosystem
contains 50 locally hosted models and 50 models accessed through APIs, with
families spanning Qwen, Llama, Mistral, DeepSeek, Gemma/Gemini, GPT, Claude,
GLM, MiniMax, and several independent releases. Table~\ref{tab:target_roster}
provides the complete roster. Each prompt is issued once to every source.
Generation uses the common settings reported in the main paper: temperature
$0.7$ and top-$p$ $0.95$.

The benchmark is balanced by construction: every source contributes exactly
500 responses and every prompt contributes exactly 100 source responses.
Source labels identify the served model endpoint, including versioned
endpoints when the provider exposes one. Each served, versioned endpoint
defines one attribution class. Family metadata is used only for grouped error
and geometry analyses.

\paragraph{Prompt composition.}
The prompts center on software-engineering and operational workflows. They
include requests to inspect, modify, test, deploy, diagnose, and document
software systems. Targets receive the prompt text through their chat
interfaces. Repository contents, tool outputs, and execution state are absent
from collection. The resulting records capture source behavior on agent-style
requests before tool execution.

Table~\ref{tab:agent500_prompt_examples} reports the complete category counts
and one recorded example per category. The category labels serve as descriptive
corpus metadata. Attribution models receive unannotated prompt--response
records. The full prompt list is part of the released
benchmark.

\begin{table*}[!t]
\centering
\scriptsize
\setlength{\tabcolsep}{4pt}
\begin{tabular*}{\textwidth}{@{\extracolsep{\fill}}p{0.28\textwidth}r p{0.60\textwidth}@{}}
\toprule
\textbf{Prompt category} & \textbf{Count} & \textbf{Representative prompt} \\
\midrule
Code modification and refactoring & 114 &
Add graceful shutdown handling to the HTTP server defined in
\texttt{cmd/server/main.go}. \\
Documentation and communication & 87 &
Draft a PR description for a change that adds rate limiting to our public API
endpoints. \\
Testing and quality assurance & 85 &
Run the failing tests in isolation and summarize what each assertion expected
versus received. \\
Code review and repository analysis & 46 &
Search the codebase for every place that hits the
\texttt{/api/payments} endpoint and list the call sites. \\
Data systems and storage & 42 &
Given a 2GB CSV of user events on disk, compute the ten most active users
without loading it all into memory. \\
Infrastructure and cloud operations & 41 &
Detect orphaned EBS volumes across all AWS accounts in our organization and
write a script for it. \\
Observability, performance, and reliability & 33 &
Our Grafana dashboard shows p99 latency spiking at 14:00 UTC daily. Which
tools would you use to correlate this with a cause? \\
Security and compliance & 24 &
I pushed a secret to git three commits ago. Walk me through the safe removal
procedure. \\
Build, CI/CD, and release engineering & 21 &
A user reports that \texttt{npm run build} fails with exit code 137. Give a
diagnosis plan before touching any files. \\
API, integration, and automation & 7 &
Translate a REST endpoint into an equivalent GraphQL query and mutation pair. \\
\bottomrule
\end{tabular*}
\caption{\textbf{Composition of the 500 Agent500 prompts.} Counts cover the
entire corpus. Examples are lightly typeset versions of prompts used during
collection. Category metadata is used only to describe the corpus.}
\label{tab:agent500_prompt_examples}
\end{table*}

\subsection{Prompt-Grouped Evaluation}

All model selection and evaluation is prompt grouped. A five-fold split with
seed 42 partitions the 500 prompt identifiers, so all 100 responses elicited
by one prompt remain together. In each outer fold, the other 400 prompts
provide 40,000 enrollment records and the held-out 100 prompts provide 10,000
single-response test records. Coordinate standardization, probe fitting, and
class-prior estimation use the enrollment records.

At budget $K>1$, test records are grouped by source into disjoint sets of $K$
responses. We use grouping seeds 42, 43, and 44 for every
$K\in\{1,5,10,20,50,100\}$. Every budget reuses the same response-level probe.
Grouping determines which held-out measurements enter the evidence sum. For
$K=1$, the grouping seeds induce the same set of 50,000 out-of-fold
predictions, making the score deterministic across seeds.

\paragraph{Metric reduction.}
The main endpoint tables report mean and standard deviation over the five
outer folds. Full budget curves pool all out-of-fold predictions before
computing each metric and then average over the three grouping seeds.
Accuracy is linear in the examples and is identical under these two
reductions. Macro-F1 is nonlinear, so pooled values can differ slightly from
the mean of fold-wise macro-F1. Macro-F1 first computes one-vs-rest F1 for
each of the 100 sources and then averages the 100 values.

\subsection{Proxy Readers and Layer Selection}

The prompt and response are jointly processed by each proxy (the \texttt{ur}
input view). The DCT receives the hidden states aligned with response tokens.
For each outer fold, its 400 enrollment prompts retain their four prompt-fold
assignments as inner folds. Every Transformer block is a candidate layer. At
each layer, we fit the coordinate standardizer and response-level source probe
on three inner folds and measure accuracy on the fourth, rotating through all
four held-out folds. We select the layer with the highest mean inner-validation
accuracy. Layer scores are compared at full numerical precision. Exact ties
favor the shallower layer. We refit the standardizer and probe at the selected
layer on all 40,000 enrollment records and evaluate once on the untouched outer
fold. The resulting fold-local layer and probe are reused for every query
budget and for all associated component ablations and shifted evaluations. The
first four proxies constitute the main roster. The supplement reports all
eight.

For the capability diagnostic in the main paper, MMLU-Pro uses
category-matched 5-shot chain-of-thought demonstrations, temperature zero,
a 4000-token output limit, and disabled thinking mode. These capability scores
serve only the proxy diagnostic.

\subsection{Shifted Evaluation Protocols}

The controlled-length test truncates Agent500 responses to the first 32 or 64
target-generation tokens before proxy extraction. The truncated versions
retain the original prompt identifiers, allowing each record to be evaluated
by the outer-fold model that excluded its prompt during training. Math100
crosses 100 mathematical problems with the same 100 candidate sources,
yielding 10,000 new records. Each of the five frozen Agent500 fold models
evaluates all Math100 responses independently. For each fold model and grouping
seed, we accumulate response-level evidence at the same six query budgets.
Tables report the mean and standard deviation of the five fold-model metrics.
Full budget curves additionally average over the three grouping seeds.

Both tests reuse the complete clean-Agent500 estimator, including its proxy
layer, coordinate standardizer, source probe, and class prior. All layer
selection, standardization, probe fitting, and prior estimation use Agent500
enrollment records exclusively. The resulting scores measure zero-retraining
transfer from the enrolled system.

\paragraph{Math100 construction.}
Math100 is a separately constructed set of 100 proof-oriented and derivational
mathematics prompts. It spans real and complex analysis, algebra, topology,
differential equations, probability, number theory, combinatorics,
optimization, numerical analysis, and functional analysis. Every source
receives the same 100 prompt texts under the Agent500 collection convention.
Evaluation scores source identity for each response.
Table~\ref{tab:math100_prompt_examples} shows representative prompts. The
domain labels provide descriptive summaries of these examples.

\begin{table*}[!t]
\centering
\scriptsize
\setlength{\tabcolsep}{4pt}
\begin{tabular*}{\textwidth}{@{\extracolsep{\fill}}p{0.20\textwidth}p{0.73\textwidth}@{}}
\toprule
\textbf{Illustrative domain} & \textbf{Representative Math100 prompt} \\
\midrule
Real analysis &
Determine, with proof, whether
$\sum_{n=2}^{\infty}[n(\ln n)^p]^{-1}$ converges as a function of $p>0$. \\
Abstract algebra &
Let $G$ be a finite group of order $p^2$, where $p$ is prime. Prove that
$G$ is abelian. \\
Topology &
Show that the fundamental group of $S^1$ is isomorphic to $\mathbb{Z}$ using
the universal cover $\mathbb{R}\to S^1$. \\
Probability &
A random walk on $\mathbb{Z}$ starts at zero and moves $+1$ with probability
$p$. Compute the probability that it ever reaches $+1$. \\
Number theory &
Prove that an integer is a sum of two squares exactly when every prime factor
congruent to $3$ modulo $4$ has even exponent. \\
Optimization &
For a strictly convex differentiable function with Lipschitz gradient, prove
that gradient descent with step size $1/L$ converges at rate $O(1/k)$. \\
\bottomrule
\end{tabular*}
\caption{\textbf{Representative prompts from Math100.} The no-retraining
domain-shift evaluation uses the complete 100-prompt collection.}
\label{tab:math100_prompt_examples}
\end{table*}

\subsection{Bench-A-Derived Pair Task}

Starting from Bench-A, introduced with MPT, we construct a balanced binary
relationship-classification task with 116 pairs spanning 67 unique models.
Positive examples pair a parent with a derived model. Negative examples pair
models from different parent families. Every model contributes 600
prompt-aligned responses, and all methods use the same 20 fixed stratified
$4{:}1$ splits. For each pair and aligned prompt, READER computes DC and
first-AC cosine similarities separately, averages each score over the 600
prompts, and trains a linear SVM on the resulting two-dimensional feature.
For each proxy, Bench-A uses the modal layer among the five Agent500 outer-fold
selections. A modal tie is resolved by mean inner-validation accuracy across
the five Agent500 outer folds and then in favor of the shallower layer.
Agent500 therefore determines the proxy layer, while Bench-A labels train only
the two-score pair readout. The two scores are standardized within each
training split. The split is applied at the pair level, and the readout is
fitted from the derived relationship labels.

The main-paper task uses in-distribution pair classification with pair-disjoint
splits, allowing individual models to recur across training and test pairs.
Two stricter protocols separate relationship transfer from recurring model
identity. In the \emph{model-disjoint} protocol, each endpoint belongs to one
side of a split. Each of 20 splits holds out a parent and at least two of its
children from each of two families. Other children from those families remain
eligible for training, exposing family-level structure while reserving the
evaluated endpoints. In the \emph{family-disjoint} protocol, each of the
$\binom{9}{2}=36$ splits holds out two complete parent families, removing both
endpoint and parent-family overlap. Positive
pairs are parent--derived edges, and negatives are sampled uniformly without
replacement from cross-family pairs. Training and test sets are balanced
independently within every split.

\section{Implementation and Baselines}
\label{app:implementation}

\subsection{READER Optimization}

Each fold-local source probe is a full-batch multinomial logistic regressor.
We standardize every fingerprint coordinate with enrollment-fold statistics,
then optimize cross-entropy with Adam for 40 steps at learning rate $10^{-3}$.
The learning rate follows cosine decay toward one percent of its initial value
over a 100-step horizon. The objective applies L2 regularization to the weight
matrix, with coefficient $\lambda=N_{\mathrm{tr}}^{-1}$ in the main-text
objective, and leaves the bias unregularized. Proxy weights are frozen.
The trainable state in each fold comprises the standardizer and
$2d\times100$ linear source head.

\subsection{Dynamic Baselines}

The dynamic baselines use the same prompt--response (\texttt{ur}) records as
the primary READER evaluation.

\paragraph{Fine-tuned DeBERTa.}
DeBERTa-v3-large is fine-tuned end to end as a 100-way text classifier under
the same five prompt-grouped folds. Each prompt--response record is serialized
as the classifier input and truncated to 256 tokenizer tokens. Each fold is
trained for two epochs with AdamW,
learning rate $6\times10^{-6}$, weight decay $0.01$, and a linear warmup ratio
of $0.06$ in FP32. At $K>1$, its response-level log-posteriors are accumulated
with the same rule as READER. This gives the text classifier the identical
test grouping and query budgets.

\paragraph{LLMmap and LLM-DNA.}
The published LLMmap and LLM-DNA evaluations rely on a controlled panel in
which candidates answer a fixed shared prompt set. We adapt both methods to
Agent500's unseen, heterogeneous prompts at interaction level. LLMmap's
closed-set behavioral score is applied to each observed prompt--response
record before query grouping.
For LLM-DNA, we freeze MPNet, BGE-large, or Qwen3-Embedding-8B, embed each
prompt--response record, and fit a record-level source probe using the same
prompt-grouped training folds. The resulting posteriors use the same evidence
accumulation and grouping as READER. This matched protocol isolates the
interaction representation under a common dynamic decision rule.

\subsection{Static Baselines}

The Bench-A-derived comparison uses each method's native static relationship
pipeline. MPT and PhyloLM operate on their pairwise behavioral features.
LLM-DNA follows its original concat--projection--centroid procedure. READER
uses the two averaged
cosine scores defined in Section~\ref{app:experimental_details}. Every method
receives the same 20 pair-level splits. Preprocessing and supervised readouts
are fitted independently within each split.

The disjoint protocols reuse these cached method-native pair features. MPT
uses first-token agreement, PhyloLM uses first-four-character agreement, and
the three LLM-DNA variants use Euclidean distance between their fixed
GRP-128 model vectors. For every method and strict split, a linear SVM and
its scalar-feature standardizer are fitted from training pairs. Thus
the strict protocols isolate identity overlap under a fixed representation and
readout family.

\FloatBarrier

%% file: appendix/target_roster.tex
\section{Agent500 Target Roster}
\label{app:target_roster}

Table~\ref{tab:target_roster} lists all 100 enrolled sources. Entries 1--50
were locally hosted checkpoints. Entries 51--100 were served API endpoints.
We preserve version suffixes because endpoint versions are distinct candidate
sources in the closed ecosystem. The roster includes closely related
variants, making within-family distinctions the hardest and most operationally
relevant part of the 100-way task.

The \emph{Access} field records the response-collection route. \emph{Family
group} provides the coarse organization used for confusion matrices and
ecosystem geometry. Exact architecture and training lineage are unannotated.

\begin{table*}[!t]
\centering
\resizebox{\textwidth}{!}{%
\begin{tabular}{@{}rlcc@{\hspace{1.2em}}rlcc@{}}
\toprule
\multicolumn{4}{c}{\textbf{Locally hosted sources}} &
\multicolumn{4}{c}{\textbf{API sources}} \\
\cmidrule(r){1-4}\cmidrule(l){5-8}
\textbf{ID} & \textbf{Model or endpoint identifier} &
\textbf{Access} & \textbf{Family group} &
\textbf{ID} & \textbf{Model or endpoint identifier} &
\textbf{Access} & \textbf{Family group} \\
\midrule
\input{appendix/generated_target_roster_paired}\end{tabular}%
}
\caption{\textbf{The 100 Agent500 candidate sources.} Model identifiers are
retained exactly as recorded in the validated experiment manifest. Access
denotes the collection route. Family groups provide coarse analysis labels.}
\label{tab:target_roster}
\end{table*}

%% file: appendix/generated_target_roster_paired.tex
% Generated by scripts/generate_aaai_target_roster.py.
% Source: /data/projects/LLM_Identifier/outputs/exp_agent_n500/100-way/targets.json
1 & \path{Qwen/Qwen2.5-3B-Instruct} & Local & Qwen & 51 & \path{gpt-4o} & API & GPT \\
2 & \path{Qwen/Qwen2.5-7B-Instruct} & Local & Qwen & 52 & \path{gpt-4o-2024-11-20} & API & GPT \\
3 & \path{Qwen/Qwen2.5-14B-Instruct} & Local & Qwen & 53 & \path{gpt-4.1-mini-2025-04-14} & API & GPT \\
4 & \path{Qwen/Qwen2.5-32B-Instruct} & Local & Qwen & 54 & \path{gpt-4.1-2025-04-14} & API & GPT \\
5 & \path{Qwen/Qwen2.5-72B-Instruct} & Local & Qwen & 55 & \path{gpt-4.1-nano-2025-04-14} & API & GPT \\
6 & \path{Qwen/Qwen3-1.7B} & Local & Qwen & 56 & \path{gpt-5.4-nano} & API & GPT \\
7 & \path{Qwen/Qwen3-1.7B-Base} & Local & Qwen & 57 & \path{qwen3.5-flash} & API & Qwen \\
8 & \path{Qwen/Qwen3-4B-Instruct-2507} & Local & Qwen & 58 & \path{qwen3.5-plus} & API & Qwen \\
9 & \path{Qwen/Qwen3-4B-Thinking-2507} & Local & Qwen & 59 & \path{gpt-5.4-mini} & API & GPT \\
10 & \path{Qwen/Qwen3-14B} & Local & Qwen & 60 & \path{gpt-5.4} & API & GPT \\
11 & \path{Qwen/Qwen3-30B-A3B-Instruct-2507} & Local & Qwen & 61 & \path{gpt-5.5} & API & GPT \\
12 & \path{Qwen/Qwen3-30B-A3B-Thinking-2507} & Local & Qwen & 62 & \path{gpt-5.6-luna} & API & GPT \\
13 & \path{Qwen/Qwen3-Next-80B-A3B-Instruct} & Local & Qwen & 63 & \path{gpt-5.6-sol} & API & GPT \\
14 & \path{Qwen/Qwen3-Next-80B-A3B-Thinking} & Local & Qwen & 64 & \path{gpt-5.6-terra} & API & GPT \\
15 & \path{Qwen/Qwen3-Coder-30B-A3B-Instruct} & Local & Qwen & 65 & \path{qwen3.6-flash} & API & Qwen \\
16 & \path{Qwen/QwQ-32B} & Local & Qwen & 66 & \path{qwen3.6-plus} & API & Qwen \\
17 & \path{Qwen/Qwen3.5-4B} & Local & Qwen & 67 & \path{glm-4.7-FlashX} & API & GLM \\
18 & \path{Qwen/Qwen3.5-4B-Base} & Local & Qwen & 68 & \path{claude-opus-4-6} & API & Claude \\
19 & \path{Qwen/Qwen3.5-9B} & Local & Qwen & 69 & \path{claude-opus-4-7} & API & Claude \\
20 & \path{Qwen/Qwen3.5-9B-Base} & Local & Qwen & 70 & \path{claude-opus-4-8} & API & Claude \\
21 & \path{Qwen/Qwen3.5-27B} & Local & Qwen & 71 & \path{claude-sonnet-4-20250514} & API & Claude \\
22 & \path{Qwen/Qwen3.5-122B-A10B} & Local & Qwen & 72 & \path{glm-4.6} & API & GLM \\
23 & \path{Qwen/Qwen3.6-27B} & Local & Qwen & 73 & \path{claude-sonnet-4-6} & API & Claude \\
24 & \path{Qwen/Qwen1.5-MoE-A2.7B-Chat} & Local & Qwen & 74 & \path{mistralai/mistral-small-24b-instruct-2501} & API & Mistral \\
25 & \path{meta-llama/Meta-Llama-3-8B-Instruct} & Local & Llama & 75 & \path{gemini-2.5-flash-lite} & API & Gemini \\
26 & \path{meta-llama/Meta-Llama-3-70B-Instruct} & Local & Llama & 76 & \path{deepseek-v4-flash} & API & DeepSeek \\
27 & \path{meta-llama/Llama-3.1-8B-Instruct} & Local & Llama & 77 & \path{qwen3.6-max-preview} & API & Qwen \\
28 & \path{meta-llama/Llama-3.1-70B-Instruct} & Local & Llama & 78 & \path{gemini-3.1-flash-lite-preview} & API & Gemini \\
29 & \path{meta-llama/Llama-3.2-1B-Instruct} & Local & Llama & 79 & \path{qwen3.7-plus} & API & Qwen \\
30 & \path{meta-llama/Llama-3.2-3B-Instruct} & Local & Llama & 80 & \path{mistralai/mistral-medium-3} & API & Mistral \\
31 & \path{meta-llama/Llama-4-Scout-17B-16E-Instruct} & Local & Llama & 81 & \path{deepseek-ai/DeepSeek-V3} & API & DeepSeek \\
32 & \path{mistralai/Mistral-7B-Instruct-v0.3} & Local & Mistral & 82 & \path{deepseek-ai/DeepSeek-V3.1-Terminus} & API & DeepSeek \\
33 & \path{mistralai/Mistral-Nemo-Instruct-2407} & Local & Mistral & 83 & \path{deepseek-ai/DeepSeek-V3.2} & API & DeepSeek \\
34 & \path{mistralai/Mixtral-8x7B-Instruct-v0.1} & Local & Mistral & 84 & \path{deepseek-v4-pro} & API & DeepSeek \\
35 & \path{deepseek-ai/DeepSeek-R1-Distill-Qwen-1.5B} & Local & DeepSeek & 85 & \path{qwen3-coder-480b-a35b-instruct} & API & Qwen \\
36 & \path{deepseek-ai/DeepSeek-R1-Distill-Qwen-7B} & Local & DeepSeek & 86 & \path{qwen3-max-2026-01-23} & API & Qwen \\
37 & \path{deepseek-ai/DeepSeek-R1-Distill-Qwen-14B} & Local & DeepSeek & 87 & \path{qwen3.5-397b-a17b} & API & Qwen \\
38 & \path{deepseek-ai/DeepSeek-R1-Distill-Qwen-32B} & Local & DeepSeek & 88 & \path{qwen3.7-max-2026-05-20} & API & Qwen \\
39 & \path{deepseek-ai/DeepSeek-V2-Lite-Chat} & Local & DeepSeek & 89 & \path{MiniMax-Text-01} & API & MiniMax \\
40 & \path{google/gemma-3-27b-it} & Local & Gemma & 90 & \path{mistralai/mistral-small-3.1-24b-instruct} & API & Mistral \\
41 & \path{google/gemma-4-26B-A4B-it} & Local & Gemma & 91 & \path{MiniMax-M3} & API & MiniMax \\
42 & \path{google/gemma-4-31B-it} & Local & Gemma & 92 & \path{glm-4.7} & API & GLM \\
43 & \path{microsoft/Phi-3-mini-4k-instruct} & Local & Phi & 93 & \path{glm-5.1} & API & GLM \\
44 & \path{tencent/Hunyuan-1.8B-Instruct} & Local & Hunyuan & 94 & \path{glm-5.2} & API & GLM \\
45 & \path{openai/gpt-oss-20b} & Local & GPT & 95 & \path{mistralai/mistral-medium-3-5} & API & Mistral \\
46 & \path{openai/gpt-oss-120b} & Local & GPT & 96 & \path{kimi-k3} & API & Kimi \\
47 & \path{zai-org/GLM-4.5-Air} & Local & GLM & 97 & \path{mistralai/mistral-large-2512} & API & Mistral \\
48 & \path{ByteDance-Seed/Seed-OSS-36B-Instruct} & Local & Seed & 98 & \path{mistralai/mistral-medium-3.1} & API & Mistral \\
49 & \path{baidu/ERNIE-4.5-21B-A3B-Thinking} & Local & ERNIE & 99 & \path{meta-llama/llama-3.3-70b-instruct} & API & Llama \\
50 & \path{LiquidAI/LFM2-24B-A2B} & Local & LFM & 100 & \path{meta-llama/llama-4-maverick} & API & Llama \\
\bottomrule

%% file: appendix/algorithm.tex
\section{READER Inference}
\label{app:algorithms}

Algorithm~\ref{alg:spectral_encoder} implements Spectral Trajectory Encoding.
Algorithm~\ref{alg:dynamic_reader} applies Bayesian Evidence Accumulation to an
arbitrary query budget using a single response-level probe.

\begin{algorithm}[t]
\caption{Spectral Trajectory Encoding}
\label{alg:spectral_encoder}
\begin{algorithmic}[1]
\REQUIRE Prompt $p$, response $x$, frozen proxy $\phi$, proxy layer $\ell$
\ENSURE Response fingerprint $\mathbf{u}\in\mathbb{R}^{2d}$
\STATE Run $\phi$ on $(p,x)$ and retain the layer-$\ell$ states aligned with
the $T$ response tokens, $\{\mathbf{h}_t\}_{t=0}^{T-1}$
\STATE $\displaystyle
\mathbf{z}_0\leftarrow
\frac{1}{T}\sum_{t=0}^{T-1}\mathbf{h}_t$
\STATE $\displaystyle
\mathbf{z}_1\leftarrow
\frac{\sqrt{2}}{T}
\sum_{t=0}^{T-1}
\mathbf{h}_t
\cos\left(\frac{\pi(t+1/2)}{T}\right)$
\STATE $\mathbf{u}\leftarrow
[\mathbf{z}_0^\top,\mathbf{z}_1^\top]^\top$
\RETURN $\mathbf{u}$
\end{algorithmic}
\end{algorithm}

\begin{algorithm}[t]
\caption{Bayesian Evidence Accumulation}
\label{alg:dynamic_reader}
\begin{algorithmic}[1]
\REQUIRE Observations $\{(p_k,x_k)\}_{k=1}^{K}$, frozen proxy $\phi$, layer
$\ell$, train-fitted standardizer $g_\psi$, source probe
$q_\theta$, class prior $\boldsymbol{\pi}$, floor $\varepsilon>0$
\ENSURE Predicted source $\widehat{y}$
\FOR{$c=1$ \TO $C$}
    \STATE $\displaystyle
    S_c\leftarrow-\frac{K-1}{K}\log\pi_c$
\ENDFOR
\FOR{$k=1$ \TO $K$}
    \STATE $\mathbf{u}_k\leftarrow
    \operatorname{SpectralEncode}(p_k,x_k;\phi,\ell)$
    \STATE $\widetilde{\mathbf{u}}_k\leftarrow g_\psi(\mathbf{u}_k)$
    \STATE $\mathbf{q}_k\leftarrow
    \operatorname{softmax}(\mathbf{W}^{\top}
    \widetilde{\mathbf{u}}_k+\mathbf{b})$
    \FOR{$c=1$ \TO $C$}
        \STATE $\displaystyle
        S_c\leftarrow S_c+
        \frac{1}{K}\log(q_{k,c}+\varepsilon)$
    \ENDFOR
\ENDFOR
\STATE $\widehat{y}\leftarrow\arg\max_{c\in\mathcal{C}}S_c$
\RETURN $\widehat{y}$
\end{algorithmic}
\end{algorithm}

\subsection{Prior Correction}

The accumulation rule follows directly from the stateless query model. For
fixed observed prompts $p_{1:K}$ and source prior $\pi_c$, conditional
independence gives
\begin{equation}
p(c\mid x_{1:K},p_{1:K})
\propto
\pi_c\prod_{k=1}^{K}p(x_k\mid p_k,c).
\label{eq:app_joint_posterior}
\end{equation}
Bayes' rule for one response gives
\begin{equation}
p(x_k\mid p_k,c)
=
\frac{p(c\mid p_k,x_k)\,p(x_k\mid p_k)}{\pi_c},
\label{eq:app_single_bayes}
\end{equation}
where prompt selection is source independent and therefore uses the same
$\pi_c$ for every request. Substituting
Eq.~\ref{eq:app_single_bayes} into Eq.~\ref{eq:app_joint_posterior}, removing
terms independent of $c$, and replacing the single-response posterior by the
probe $q_\theta$ yields
\begin{equation}
\log p(c\mid x_{1:K},p_{1:K})
\doteq
\sum_{k=1}^{K}\log q_\theta(c\mid\mathbf{u}_k)
-(K-1)\log\pi_c,
\label{eq:app_prior_corrected_sum}
\end{equation}
where $\doteq$ denotes equality up to a class-independent constant. Dividing
by $K$ preserves the ranking and produces the score used in the paper. This
normalization keeps score magnitudes comparable across query budgets. In the
balanced Agent500 ecosystem, $\pi_c=1/C$, so the correction is constant and
the classifier reduces to mean log-posterior accumulation.
The $1/K$ normalization preserves class rankings while leaving the softmax
scale budget dependent. We use the accumulated outputs as decision scores.
Budget-specific held-out calibration is required for a probabilistic
interpretation.

The factorization applies to request state and permits prompts from different
tasks and domains. It is exact for independent, stateless generations
conditioned on their source and observed prompts. Shared hidden conversational
state turns Eq.~\ref{eq:app_prior_corrected_sum} into a composite evidence
score. A state-aware model would be required to calibrate that setting.
Agent500 uses independent requests.

\subsection{Dimensions and Cost}

For proxy hidden size $d$, one response produces two $d$-dimensional
coefficients and a $2d$-dimensional fingerprint. Computing both coefficients
requires one pass over the $T$ response states, $O(Td)$ arithmetic, and can be
implemented without materializing a $T\times T$ transform. The dense
$C$-source probe costs $O(Cd)$ per response, and accumulation costs $O(KC)$
for a group of $K$ observations. Stored fingerprints require $2d$ values per
record. Once the target response is observed, the audit invokes the frozen
proxy, which dominates its computational cost.

\FloatBarrier

%% file: appendix/full_results.tex
\section{Complete Attribution Results}
\label{app:full_results}

\subsection{Fold-Wise Endpoint Variation}

Table~\ref{tab:agent500_full_std} aligns the endpoint comparison with online
cost. Its standard deviations measure sensitivity to the composition of each
held-out 100-prompt group. Across all eight proxy readers, fold variation is
small relative to READER's high-budget margin over the dynamic baselines.

\begin{table*}[t]
\centering
\scriptsize
\setlength{\tabcolsep}{2.1pt}
\begin{tabular*}{\textwidth}{@{\extracolsep{\fill}}lrrrrrrrr@{}}
\toprule
\textbf{Method / backbone} &
\multicolumn{2}{c}{\textbf{Attribution at $K=1$}} &
\multicolumn{2}{c}{\textbf{Attribution at $K=100$}} &
\multicolumn{2}{c}{\textbf{Offline fit}} &
\multicolumn{2}{c}{\textbf{Online audit}} \\
\cmidrule(lr){2-3}\cmidrule(lr){4-5}
\cmidrule(lr){6-7}\cmidrule(lr){8-9}
{} &
\textbf{Acc.} & \textbf{M-F1} &
\textbf{Acc.} & \textbf{M-F1} &
\shortstack{\textbf{Trainable}\\\textbf{params.}} &
\shortstack{\textbf{Fit time}\\\textbf{per fold}} &
\shortstack{\textbf{Audit-path}\\\textbf{params.}} &
\shortstack{\textbf{Latency}\\\textbf{(s/resp.)}} \\
\midrule
Random &
 .010$\pm$.001 & .010$\pm$.001 & .010$\pm$.010 & .007$\pm$.008 &
 -- & -- & -- & -- \\
FT DeBERTa / DeBERTa-v3-L &
 .330$\pm$.011 & .295$\pm$.011 & .568$\pm$.029 & .471$\pm$.030 &
 435.2M & 7.3 min & 435.2M & .00290 \\
LLMmap\textsuperscript{\dag} / E5-L &
 .212$\pm$.003 & .189$\pm$.005 & .522$\pm$.040 & .423$\pm$.039 &
 3.05M & 10.3 s & 562.9M & .00061 \\
DNA\textsuperscript{\dag} / MPNet &
 .119$\pm$.004 & .109$\pm$.005 & .706$\pm$.043 & .644$\pm$.045 &
 76.9K & 0.17 s & 109.6M & .00101 \\
DNA\textsuperscript{\dag} / BGE-L &
 .191$\pm$.005 & .188$\pm$.005 & .790$\pm$.028 & .739$\pm$.034 &
 102.5K & 0.17 s & 335.2M & .00291 \\
DNA\textsuperscript{\dag} / Qwen3-Emb-8B &
 .185$\pm$.006 & .192$\pm$.004 & .786$\pm$.012 & .735$\pm$.016 &
 409.7K & 0.41 s & 7.57B & .00613 \\
\addlinespace[2pt]
FT DNA\textsuperscript{\dag} / MPNet &
 .230$\pm$.008 & .170$\pm$.010 & .310$\pm$.023 & .218$\pm$.022 &
 109.6M & 2.4 min & 109.6M & .00101 \\
FT DNA\textsuperscript{\dag} / BGE-L &
 .303$\pm$.008 & .237$\pm$.009 & .458$\pm$.020 & .347$\pm$.021 &
 335.2M & 5.3 min & 335.2M & .00291 \\
FT DNA\textsuperscript{\dag} / Qwen3-Emb-8B &
 \best{.554$\pm$.005} & \best{.530$\pm$.005} &
 .850$\pm$.017 & .806$\pm$.022 &
 44.1M & 173 min & 7.61B & .00613 \\
\midrule
READER / Llama-3.1-8B-Instruct &
 .456$\pm$.011 & .455$\pm$.011 & .948$\pm$.013 & .931$\pm$.018 &
 819.3K & 0.55 s & 962.4M & .00053 \\
READER / Gemma-4-12B &
 .460$\pm$.012 & .457$\pm$.012 &
 .958$\pm$.015 & .945$\pm$.019 &
 768.1K & 0.63 s & 2.13B & .00123 \\
READER / Ministral-3-8B &
 .483$\pm$.011 & .481$\pm$.011 &
 .952$\pm$.017 & .939$\pm$.021 &
 819.3K & 0.64 s & 4.25B & .00274 \\
READER / Qwen3.5-9B &
 .504$\pm$.011 & .503$\pm$.010 &
 \second{.962$\pm$.024} & \second{.952$\pm$.029} &
 819.3K & 0.35 s & 4.91B & .00650 \\
\addlinespace[2pt]
READER / OLMo-3-7B-Instruct &
 .463$\pm$.009 & .462$\pm$.009 & .942$\pm$.016 & .930$\pm$.019 &
 819.3K & 0.35 s & 3.25B & .00247 \\
READER / Qwen3-8B &
 .447$\pm$.010 & .447$\pm$.010 & .948$\pm$.024 & .933$\pm$.029 &
 819.3K & 0.36 s & 1.01B & .00100 \\
READER / Qwen3-32B &
 .509$\pm$.011 & .508$\pm$.011 &
 \best{.964$\pm$.021} & \best{.955$\pm$.024} &
 1.02M & 0.43 s & 24.18B & .02141 \\
READER / Qwen3.5-27B &
 \second{.523$\pm$.009} & \second{.522$\pm$.009} &
 \second{.962$\pm$.007} & \second{.952$\pm$.008} &
 1.02M & 0.42 s & 19.92B & .07903 \\
\bottomrule
\end{tabular*}
\caption{\textbf{Agent500 attribution and computational
profile.} Results are mean$\pm$standard deviation over five prompt-grouped
folds. Computational measurements use one NVIDIA H200 with 140 GB HBM.
Trainable parameters and fit time cover one 40,000-response outer-training
fold. For READER, fit time starts from cached fingerprints for every candidate
layer and includes all inner-fold probe fits, layer selection, and the final
outer-training probe fit. Proxy forward passes are reported through audit-time
latency.
Audit-path parameters include the representation path and 100-way head.
Latency is amortized per response at each implementation's validated batch
size. FT DNA updates the complete MPNet/BGE encoder and uses LoRA for
Qwen3-Embedding-8B. \textsuperscript{\dag}Dynamic adaptation of a fixed-panel
method.}
\label{tab:agent500_full_std}
\label{tab:proxy_roster}
\label{tab:compute_profile}
\end{table*}
% Sources:
% outputs/exp_agent_n500/100-way/dct_alltokens_fullrank_best_layer_v2/reports/agent500/
% outputs/exp_agent_n500/100-way/finetuned_text_classifier/deberta_v3_large_generalization_v2/
% outputs/exp_agent_n500/100-way/llmmap_closed_q1_bea_v1/
% outputs/exp_agent_n500/100-way/reports_agg_logposterior/
% outputs/exp_agent_n500/100-way/finetuned_sentence_encoder/
% outputs/exp_agent_n500/100-way/paper_artifacts/fullrank/compute_cost_v1/
% outputs/exp_agent_n500/100-way/aaai_fold_std_v1/

FT DNA with Qwen3-Emb leads at $K=1$ ($.554$ versus $.504$). At $K=100$,
READER with Qwen3.5-9B reaches $.962$ accuracy and $.952$ macro-F1, reversing
the ranking by $11.2$ and $14.6$ points. READER uses a
$54\times$ smaller trainable set ($819.3$K versus $44.1$M) and a
$1.55\times$ smaller online path ($4.91$B versus $7.61$B). With cached
fingerprints, fitting is roughly $30{,}000\times$ shorter ($0.35$ seconds
versus $173$ minutes), with comparable runtime ($.00650$ versus $.00613$
seconds per response). FT adaptation favors single-shot classification.
READER combines lower adaptation cost with substantially stronger cumulative
attribution.

\subsection{All Query Budgets and Proxy Readers}

\begin{figure*}[!t]
  \centering
  \includegraphics[width=0.985\textwidth]{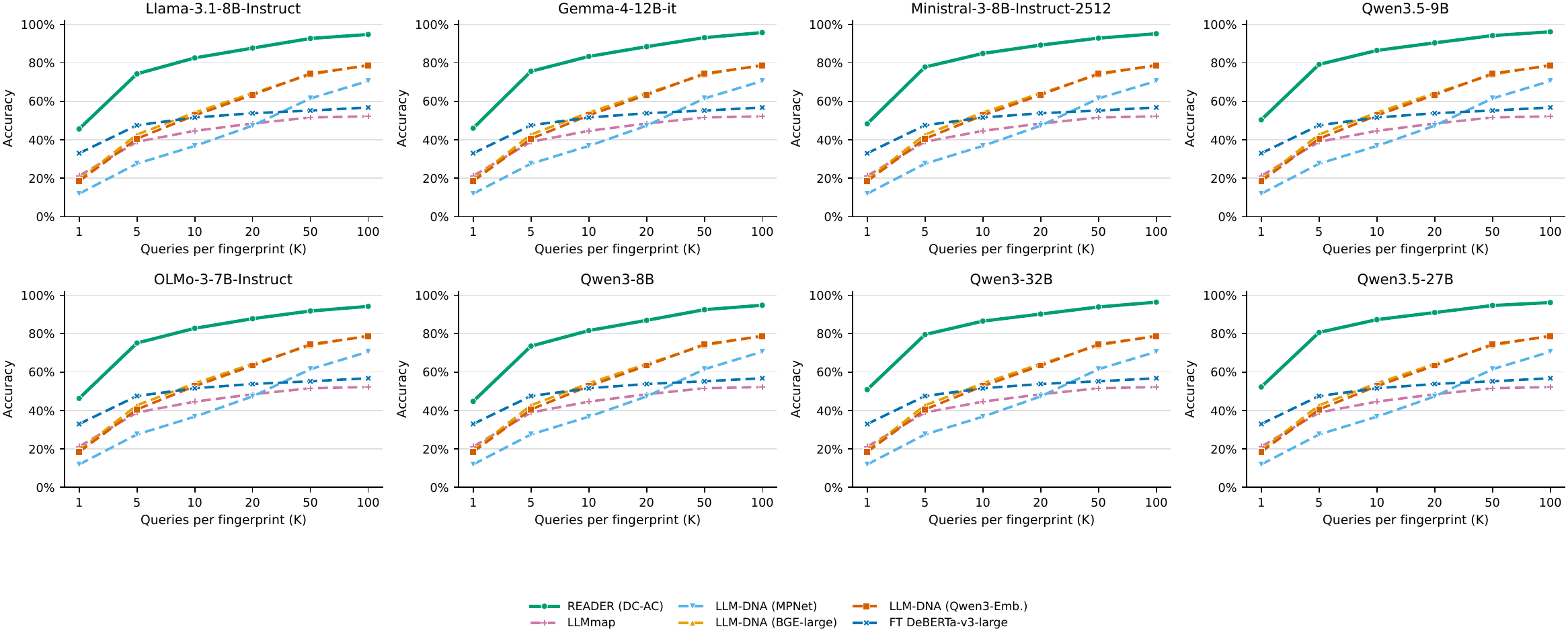}
  \par\vspace{1mm}
  \includegraphics[width=0.985\textwidth]{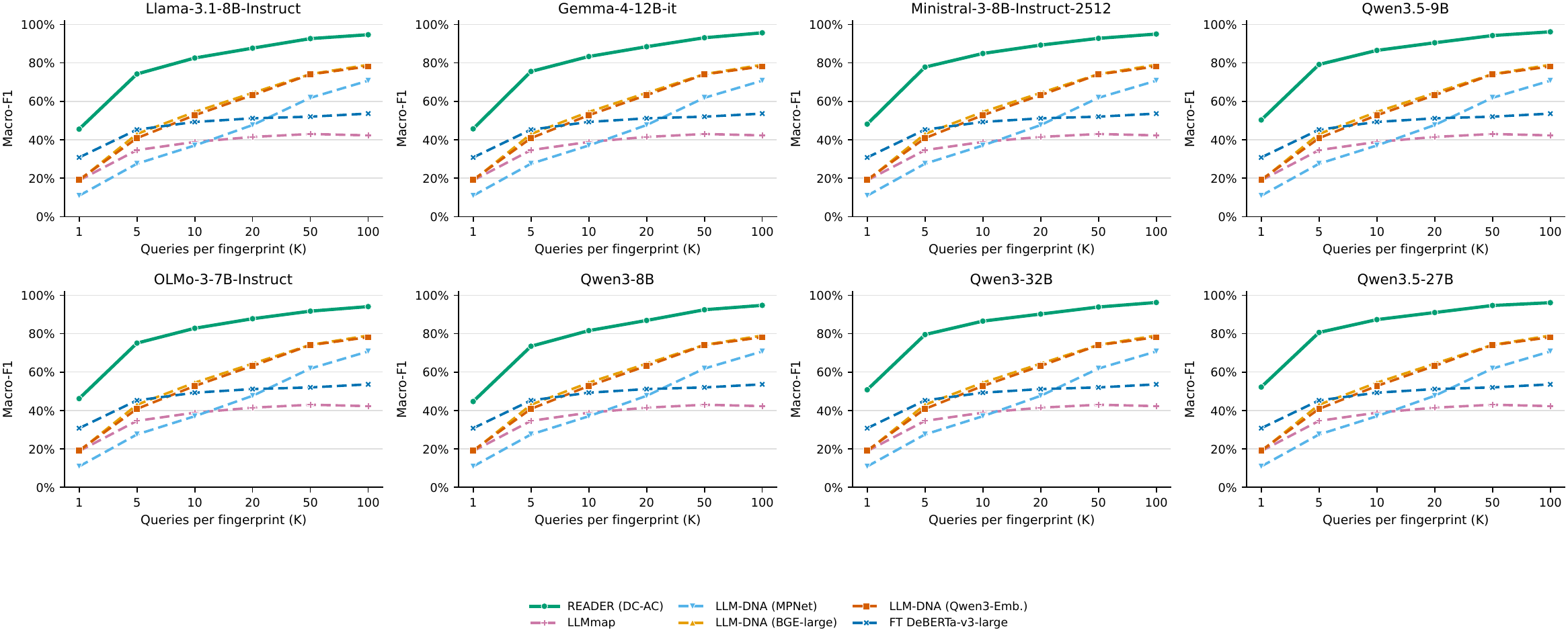}
  \caption{\textbf{Complete Agent500 accuracy (top) and macro-F1 (bottom)
  curves.} Within each outer fold, the enrollment-fitted response-level probe
  and selected layer are reused across budgets. Baselines use the same held-out
  prompt groups and evidence budgets.
  Macro-F1 is computed after pooling the out-of-fold predictions for each
  grouping seed and therefore differs slightly from the mean of fold-wise
  macro-F1 in Table~\ref{tab:agent500_full_std}.}
  \label{fig:agent500_accuracy_full}
\end{figure*}

Table~\ref{tab:agent500_allk} reports the complete accuracy trajectory for all
eight frozen proxies. Figure~\ref{fig:agent500_accuracy_full} places the
dynamic baselines on the same axes and gives the corresponding macro-F1
curves.
Every reader improves monotonically with additional responses. The spread
between proxies is modest relative to the margin over the baselines: at
$K=100$, all eight readers lie between $94.2\%$ and $96.4\%$ accuracy.
Qwen3.5-27B gives the strongest single-response result at $52.3\%$.
Qwen3-32B attains the highest Acc.@100 at $96.4\%$.

\begin{table*}[!t]
\centering
\small
\setlength{\tabcolsep}{4.5pt}
\begin{tabular*}{\textwidth}{@{\extracolsep{\fill}}lrrrrrr@{}}
\toprule
\textbf{Proxy} & \textbf{$K=1$} & \textbf{$K=5$} & \textbf{$K=10$} &
\textbf{$K=20$} & \textbf{$K=50$} & \textbf{$K=100$} \\
\midrule
Llama-3.1-8B-Instruct & .456 & .743 & .826 & .877 & .927 & .948 \\
Gemma-4-12B-it        & .460 & .756 & .834 & .885 & .932 & .958 \\
Ministral-3-8B        & .483 & .779 & .849 & .893 & .929 & .952 \\
Qwen3.5-9B            & .504 & .792 & .865 & .905 & .942 & .962 \\
\midrule
OLMo-3-7B-Instruct    & .463 & .751 & .828 & .878 & .918 & .942 \\
Qwen3-8B              & .447 & .735 & .816 & .869 & .925 & .948 \\
Qwen3-32B             & .509 & .795 & .865 & .902 & .939 & \best{.964} \\
Qwen3.5-27B           & \best{.523} & \best{.806} & \best{.873} &
\best{.910} & \best{.947} & .962 \\
\bottomrule
\end{tabular*}
\caption{\textbf{Agent500 accuracy across all query budgets and proxies.}
Values pool out-of-fold predictions and average over three grouping seeds.
The first four rows are the main proxy roster.}
\label{tab:agent500_allk}
\end{table*}

\subsection{Capability Correlations Across Query Budgets}
\label{app:capability_correlations}

\begin{figure*}[t]
  \centering
  \begin{minipage}[t]{0.485\textwidth}
    \centering
    \includegraphics[width=\linewidth]{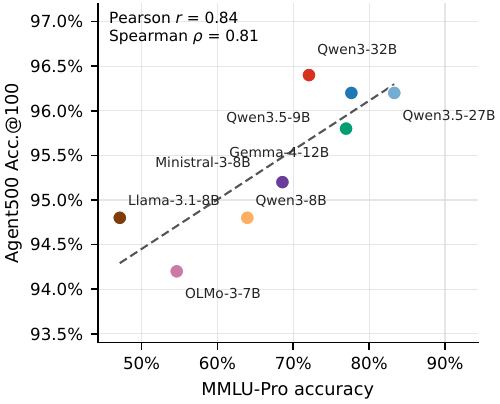}
  \end{minipage}
  \hfill
  \begin{minipage}[t]{0.485\textwidth}
    \centering
    \includegraphics[width=\linewidth]{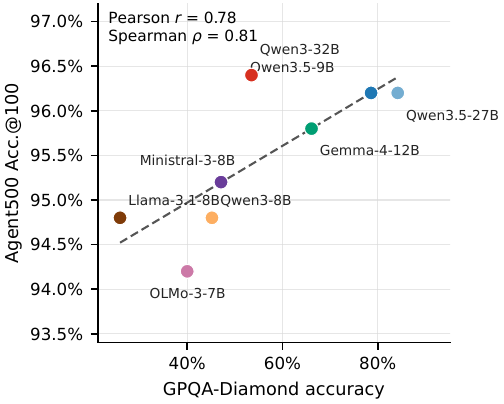}
  \end{minipage}
  \caption{\textbf{Proxy capability and attribution at $K=100$.}
  The left panel uses our MMLU-Pro scores. The right uses non-reasoning
  GPQA-Diamond scores from Artificial Analysis. Agent500 accuracy is measured
  by us, and dashed lines are least-squares fits over the same eight frozen
  proxies.}
  \label{fig:capability_full_budget}
\end{figure*}

The main paper relates proxy capability to attribution after evidence
accumulation. The left panel of Figure~\ref{fig:capability_full_budget}
reports this diagnostic using MMLU-Pro, which we evaluate under the
non-thinking 5-shot protocol specified in
Section~\ref{app:experimental_details}. At $K=100$, the association is strong
in both value and rank (Pearson $r=.84$, Spearman $\rho=.81$).

We test whether this pattern depends on one capability benchmark using
GPQA-Diamond scores independently reported by Artificial Analysis.\footnote{
\url{https://artificialanalysis.ai/evaluations/gpqa-diamond}} All eight scores
use the non-reasoning setting. The right panel yields Pearson $r=.78$ and
Spearman $\rho=.81$ at $K=100$. Across this eight-proxy roster, MMLU-Pro and
GPQA-Diamond yield the same rank ordering and similar positive associations
with Acc.@100. We treat this as a small-roster diagnostic of proxy choice.

\begin{table}[H]
\centering
\footnotesize
\setlength{\tabcolsep}{3.5pt}
\begin{tabular*}{\columnwidth}{@{\extracolsep{\fill}}lccc@{}}
\toprule
\textbf{Budget} & \textbf{MMLU $r$} & \textbf{GPQA $r$} &
\textbf{Shared $\rho$} \\
\midrule
$K=1$   & .695 (.056) & .713 (.047) & .714 (.047) \\
$K=100$ & .838 (.009) & .779 (.023) & .807 (.015) \\
\bottomrule
\end{tabular*}
\caption{\textbf{Capability--attribution correlations at the endpoint
budgets.} Parentheses contain two-sided $p$-values over the eight proxies.
MMLU-Pro and GPQA-Diamond induce the same proxy ranking, so their Spearman
correlations are identical at a fixed $K$.}
\label{tab:capability_budget_correlations}
\end{table}

Table~\ref{tab:capability_budget_correlations} summarizes both endpoint
budgets. The positive association appears at $K=1$ and is stronger at
$K=100$ within this roster. Pearson correlation retains benchmark-specific
score spacing. Spearman correlation is shared because the two benchmarks
induce the same proxy ordering.

\subsection{Residual Confusion}

\begin{figure*}[t]
  \centering
  \includegraphics[width=\textwidth]{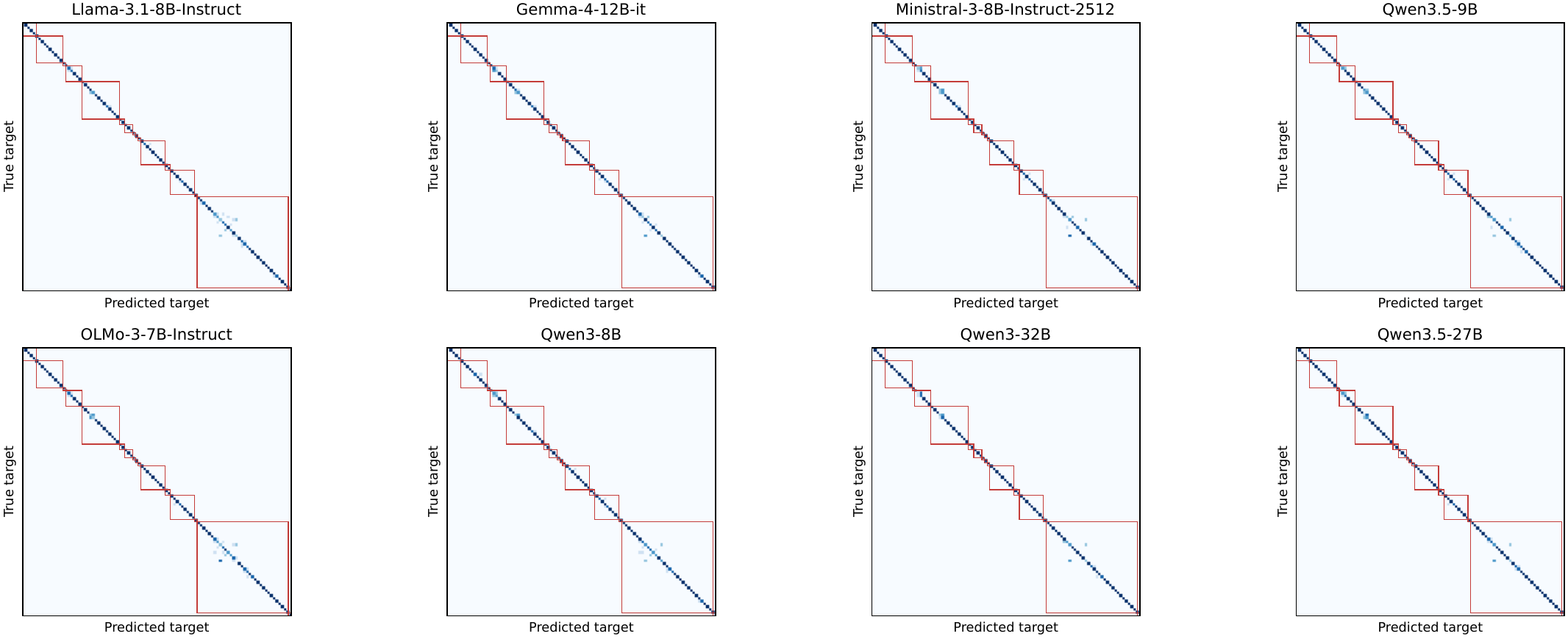}
  \caption{\textbf{Out-of-fold confusion at $K=100$ for all proxy readers.}
  Matrices are row normalized. Red boxes delimit target families.}
  \label{fig:confusion_full}
\end{figure*}

Figure~\ref{fig:confusion_full} expands the main-paper confusion analysis to
all eight proxies. Targets are ordered by provider family. At $K=100$, errors
remain tightly concentrated in the diagonal family blocks across both the
main and supplementary readers. The eight readers show qualitatively similar
within-family confusion patterns. Closely related endpoint releases and
instruction variants account for the residual errors.

\begin{table*}[!t]
\centering
\small
\setlength{\tabcolsep}{4pt}
\begin{tabular*}{\textwidth}{@{\extracolsep{\fill}}lcccccc@{}}
\toprule
& \multicolumn{2}{c}{\textbf{32 tokens}} &
  \multicolumn{2}{c}{\textbf{64 tokens}} &
  \multicolumn{2}{c}{\textbf{Math100}} \\
\cmidrule(lr){2-3}\cmidrule(lr){4-5}\cmidrule(l){6-7}
\textbf{Method} & \textbf{Acc.} & \textbf{M-F1} &
\textbf{Acc.} & \textbf{M-F1} &
\textbf{Acc.} & \textbf{M-F1} \\
\midrule
Random                                  & -- & -- & -- & -- &
  .010$\pm$.010 & .007$\pm$.008 \\
LLMmap\textsuperscript{\dag}            & .058 & .046 & .144 & .109 &
  .128$\pm$.037 & .065$\pm$.029 \\
DNA / MPNet\textsuperscript{\dag}       & .034 & .029 & .026 & .018 &
  .046$\pm$.008 & .018$\pm$.004 \\
DNA / BGE\textsuperscript{\dag}         & .026 & .011 & .038 & .032 &
  .102$\pm$.025 & .055$\pm$.017 \\
DNA / Qwen-Emb.\textsuperscript{\dag}   & .014 & .006 & .020 & .012 &
  .142$\pm$.016 & .077$\pm$.008 \\
FT DeBERTa                              & .158 & \best{.143} & .278 & .259 &
  .232$\pm$.018 & .151$\pm$.016 \\
\midrule
READER / Llama                          & \best{.185} & \second{.142} &
  .407 & .343 & .262$\pm$.022 & .170$\pm$.019 \\
READER / Gemma                          & \second{.170} & .132 &
  \second{.429} & \second{.379} & .204$\pm$.005 & .127$\pm$.008 \\
READER / Ministral                      & .138 & .095 & .368 & .307 &
  \second{.282$\pm$.012} & \second{.191$\pm$.013} \\
READER / Qwen                           & .159 & .111 &
  \best{.505} & \best{.439} &
  \best{.330$\pm$.028} & \best{.242$\pm$.025} \\
\bottomrule
\end{tabular*}
\caption{\textbf{No-retraining stress tests at $K=100$.}
The first two column groups evaluate controlled response truncation. The final
group transfers the Agent500 fold models to Math100 and reports
mean$\pm$standard deviation across five folds.
\textsuperscript{\dag}Dynamic adaptation of a fixed-panel method.}
\label{tab:stress_tests_full}
\end{table*}

\subsection{No-Retraining Stress Tests}
\label{app:stress_tests}

The following tests hold the complete Agent500 enrollment pipeline fixed while
varying the observed responses. Every method reuses its clean fold model,
including all preprocessing and source-head parameters.

\subsubsection{Controlled Response Length}

The response-length columns of Table~\ref{tab:stress_tests_full} expand the
endpoint comparison across the four main proxies and matched baselines.
Truncation causes a substantial absolute decline, particularly at 32 tokens.
At 64 tokens, READER-Qwen reaches $50.5\%$ accuracy and READER-Gemma reaches
$42.9\%$, both above fine-tuned DeBERTa's $27.8\%$. At 32 tokens, READER and
DeBERTa are much closer. Severe shortening shifts the decision boundary
learned from full responses, locating a clear response-coverage limit.

\subsubsection{Math100 Domain Shift}
\label{app:math_full}

\begin{figure*}[t]
  \centering
  \includegraphics[width=\textwidth]{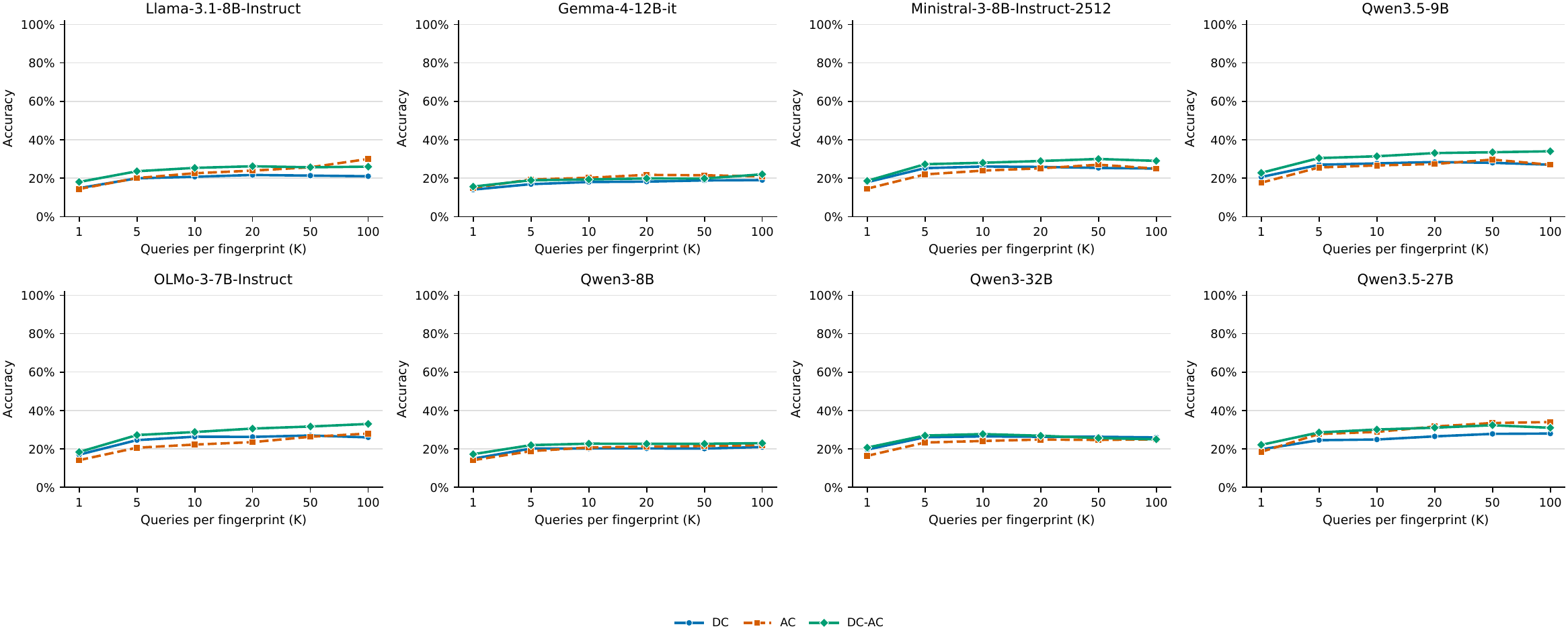}
  \par\medskip
  \includegraphics[width=\textwidth]{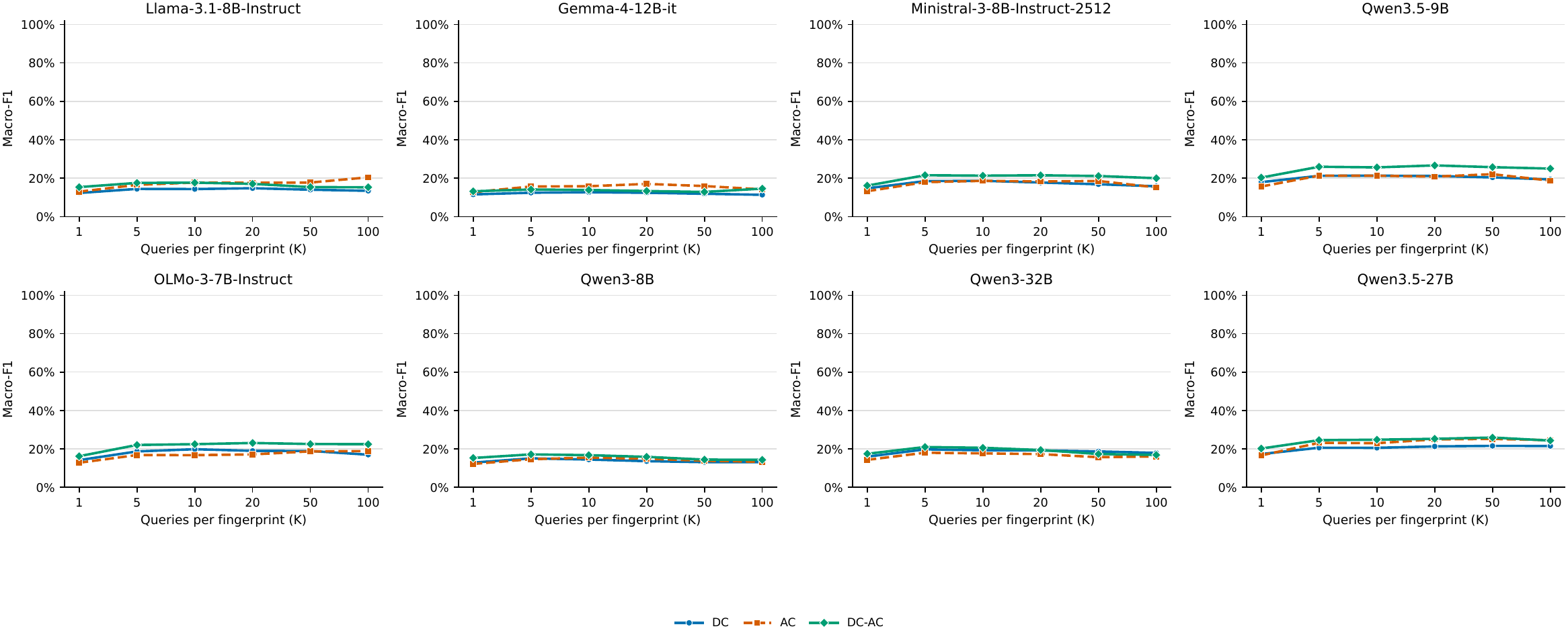}
  \caption{\textbf{No-retraining Math100 accuracy (top) and macro-F1
  (bottom).} Curves average metrics across the five independently evaluated
  Agent500 fold models and three grouping seeds. Domain shift depresses
  macro-F1 more strongly than accuracy, indicating uneven transfer across the
  100 candidate sources.}
  \label{fig:math_accuracy_full}
\end{figure*}

The Math100 columns of Table~\ref{tab:stress_tests_full} give the five-fold
variation at $K=100$. Proxy choice has a larger effect under this domain
shift. READER-Qwen and READER-Ministral reach $33.0\%$ and $28.2\%$ accuracy,
respectively, whereas READER-Gemma reaches $20.4\%$.
The same separation appears across query budgets in
Figure~\ref{fig:math_accuracy_full}. Additional evidence improves both
metrics for every reader. The shifted-domain ceiling remains well below
in-domain performance, highlighting the value of enrollment coverage over the
target prompt distribution.
% Sources:
% outputs/exp_agent_n500/100-way/dct_alltokens_fullrank_best_layer_v2/reports/generalization_ur/
% outputs/exp_agent_n500/100-way/finetuned_text_classifier/deberta_v3_large_generalization_v2/
% outputs/exp_agent_n500/100-way/llmmap_closed_q1_bea_v1/
% outputs/exp_agent_n500/100-way/reports_agg_logposterior/

\begin{table*}[!t]
\centering
\small
\setlength{\tabcolsep}{5pt}
\begin{tabular*}{\textwidth}{@{\extracolsep{\fill}}lccccc@{}}
\toprule
\textbf{Method} & \textbf{Acc.} & \textbf{Precision} & \textbf{Recall} &
\textbf{F1} & \textbf{AUC} \\
\midrule
MPT                    & .742$\pm$.061 & \second{.943$\pm$.082} &
                         .521$\pm$.123 & .660$\pm$.105 & .746$\pm$.085 \\
PhyloLM                & .758$\pm$.057 & \best{.944$\pm$.073} &
                         .554$\pm$.113 & .690$\pm$.093 & .793$\pm$.078 \\
DNA / MPNet            & .621$\pm$.063 & .650$\pm$.091 &
                         .554$\pm$.119 & .589$\pm$.080 & .675$\pm$.074 \\
DNA / BGE              & .631$\pm$.076 & .629$\pm$.078 &
                         \best{.646$\pm$.139} & .631$\pm$.096 & .722$\pm$.075 \\
DNA / Qwen-Emb.        & .704$\pm$.090 & .835$\pm$.162 &
                         .554$\pm$.116 & .650$\pm$.096 & .723$\pm$.086 \\
\midrule
READER / Llama         & \best{.781$\pm$.059} & .928$\pm$.079 &
                         .617$\pm$.127 & \best{.731$\pm$.093} &
                         \best{.828$\pm$.065} \\
READER / Gemma         & .694$\pm$.068 & .730$\pm$.091 &
                         \second{.629$\pm$.125} & .668$\pm$.086 &
                         .768$\pm$.078 \\
READER / Ministral     & .754$\pm$.063 & .906$\pm$.089 &
                         .575$\pm$.131 & .693$\pm$.099 &
                         \second{.811$\pm$.063} \\
READER / Qwen          & \second{.767$\pm$.062} & .907$\pm$.088 &
                         .600$\pm$.122 & \second{.714$\pm$.096} &
                         .809$\pm$.074 \\
\bottomrule
\end{tabular*}
\caption{\textbf{Static relationship classification derived from Bench-A
with split variation.} READER uses the DC--AC two-score readout.
Mean$\pm$standard deviation over 20 fixed stratified splits.}
\label{tab:bench_a_full_std}
\end{table*}

\begin{table*}[!t]
\centering
\scriptsize
\setlength{\tabcolsep}{2.5pt}
\begin{tabular*}{\textwidth}{@{\extracolsep{\fill}}lccc@{\quad}ccc@{}}
\toprule
& \multicolumn{3}{c}{\textbf{Model-disjoint, family visible}} &
  \multicolumn{3}{c}{\textbf{Family-disjoint}} \\
\cmidrule(lr){2-4}\cmidrule(lr){5-7}
\textbf{Method} &
\textbf{Acc. [95\% CI]} & \textbf{F1} & \textbf{AUC} &
\textbf{Acc. [95\% CI]} & \textbf{F1} & \textbf{AUC} \\
\midrule
MPT &
.622 [.571,.675] & .611 & \second{.909} &
.677 [.639,.716] & .674 & \second{.901} \\
PhyloLM &
.664 [.615,.715] & .623 & \best{.913} &
\second{.706 [.661,.753]} & \second{.683} & \best{.904} \\
DNA / MPNet &
.613 [.563,.665] & .589 & .839 &
.630 [.597,.663] & .617 & .806 \\
DNA / BGE &
.645 [.554,.735] & .673 & .803 &
.670 [.632,.707] & .678 & .803 \\
DNA / Qwen-Emb. &
.655 [.598,.712] & .612 & .903 &
.667 [.636,.698] & .640 & .873 \\
\midrule
READER / Llama &
\best{.756 [.689,.822]} & \best{.735} & .895 &
.663 [.619,.707] & .669 & .825 \\
READER / Gemma &
.657 [.581,.732] & .671 & .854 &
.624 [.591,.656] & .646 & .772 \\
READER / Ministral &
.699 [.620,.778] & .664 & .864 &
.663 [.626,.701] & .649 & .813 \\
READER / Qwen &
\second{.751 [.676,.821]} & \second{.697} & .856 &
\best{.732 [.693,.771]} & \best{.685} & .830 \\
\bottomrule
\end{tabular*}
\caption{\textbf{Bench-A-derived relationship transfer under identity
holdout.} Means are over 20 model-disjoint or 36 leave-two-family-out
splits. Accuracy intervals use a percentile bootstrap over splits. Since
held-out sets overlap across splits, the intervals and paired tests quantify
sensitivity across the specified protocols.}
\label{tab:bench_a_disjoint}
\end{table*}

\clearpage

\raggedbottom

\section{Complete Static Relationship Results}
\label{app:static_full}

Table~\ref{tab:bench_a_full_std} reports variation over the 20 fixed
pair-level splits of the balanced task derived from Bench-A. The relatively
large standard deviations reflect the small test set of 24 pairs per split.
READER-Llama has the best mean accuracy, F1, and AUC. PhyloLM has the best
precision.

% Sources:
% outputs/bench_a_pairwise_blackbox_116/dct_component_grp128_comparison/
% outputs/bench_a_pairwise_blackbox_116/reports/

\subsection{Transfer to Unseen Models and Families}
\label{app:bench_a_disjoint}

Table~\ref{tab:bench_a_disjoint} separates pair classification from recurring
endpoint identity. Removing model overlap lowers every READER result relative
to the in-distribution pair splits. Even under model holdout, Llama and Qwen
reach $75.6\%$ and $75.1\%$ accuracy, exceeding PhyloLM by 9.2 and 8.7
points. The corresponding intervals remain wide because each split contains
few positive parent--derived edges.

Holding out complete parent families is the harder boundary. Qwen is strongest
at $73.2\%$ accuracy. The other readers fall below PhyloLM. Among the paired
comparisons with PhyloLM, Gemma is the only difference that remains significant
after Holm correction. It is 8.2 points lower under family holdout
($p_{\mathrm{Holm}}=.012$). Relationship transfer therefore depends strongly
on the held-out parent family and proxy reader.

% Sources:
% outputs/bench_a_pairwise_blackbox_116/disjoint_protocol_v1/
% outputs/bench_a_pairwise_blackbox_116/disjoint_evaluation_v1/

The spectral decomposition also clarifies family-transfer behavior.
Under family holdout, first AC alone exceeds the joint DC--AC readout by
11.1, 8.9, 9.1, and 1.5 accuracy points for Llama, Gemma, Ministral, and
Qwen, respectively. DC carries most of the dynamic source signal. In static
family transfer, adding the DC score lowers accuracy by the reported amounts.
First AC therefore provides the stronger family-transfer readout, while the
joint fingerprint remains the default for the primary dynamic task.

\begin{table}[H]
\centering
\small
\setlength{\tabcolsep}{5pt}
\begin{tabular*}{\columnwidth}{@{\extracolsep{\fill}}lrrr@{}}
\toprule
\textbf{Pair feature} & \textbf{Acc.} & \textbf{F1} & \textbf{AUC} \\
\midrule
DC cosine                 & .541 & .636 & .621 \\
First-AC cosine           & \best{.749} & \best{.709} & .800 \\
DC--AC score readout      & \best{.749} & .702 & \best{.804} \\
GRP-128 first-AC cosine   & .747 & .701 & .797 \\
\bottomrule
\end{tabular*}
\caption{\textbf{Spectral components on the Bench-A-derived pair task.}
Four-proxy means over 20 fixed splits. The last row applies a data-independent
Gaussian random projection before cosine similarity.}
\label{tab:component_static_detail}
\end{table}

%% file: appendix/ablations_and_boundaries.tex
\section{Representation Ablations and Diagnostics}
\label{app:ablations}

\subsection{DC and First-AC Components}

\begin{figure*}[t]
  \centering
  \includegraphics[width=\textwidth]{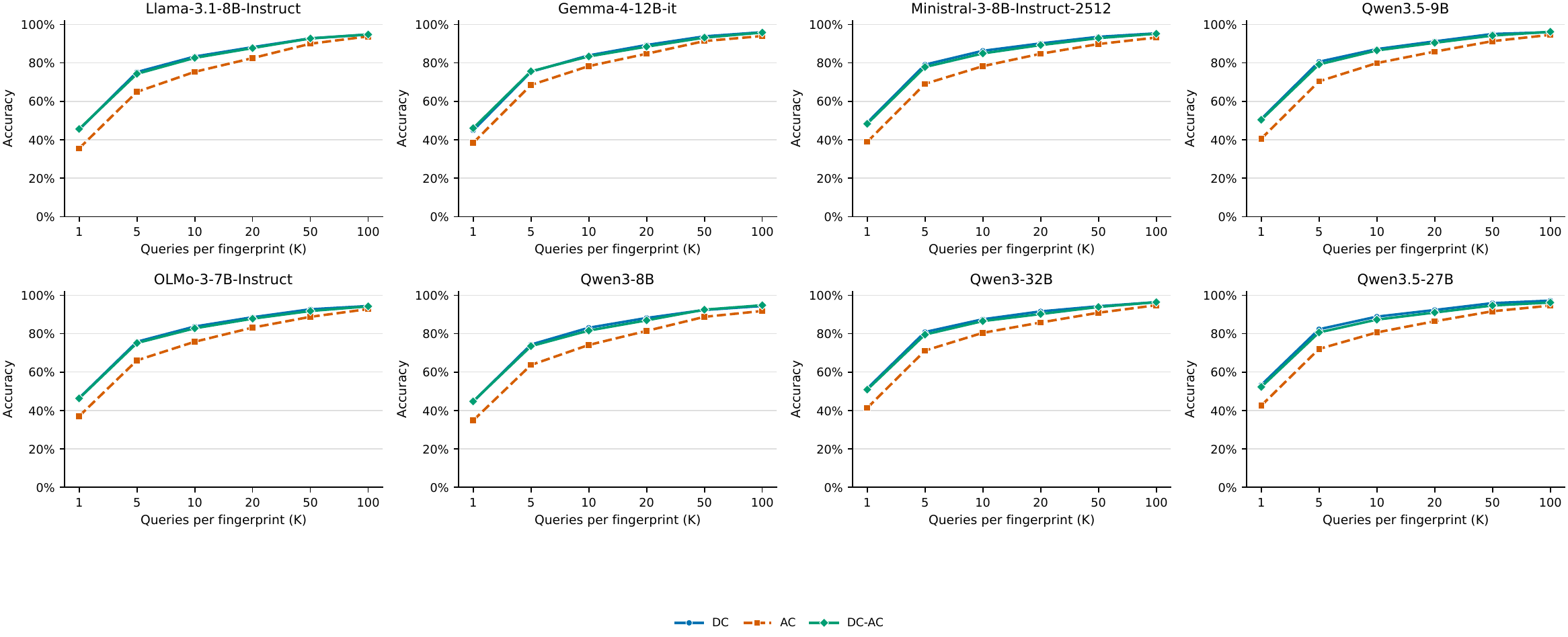}
  \caption{\textbf{Spectral-component accuracy on Agent500.} DC, first AC,
  and their full-rank concatenation use independently fitted probes at the
  same fold-local selected proxy layer.}
  \label{fig:component_accuracy_full}
\end{figure*}

Figure~\ref{fig:component_accuracy_full} gives the complete Agent500
component ablation. DC carries most of the dynamic source signal for every
proxy and budget. First AC alone is consistently weaker, especially from one
response, yet it remains far above chance and approaches DC as evidence
accumulates. The joint DC--AC fingerprint either matches DC or changes it by
at most a small amount. Its consistency across eight readers extends the central
dynamic pattern beyond the four-model main roster.

The static relationship task reverses the component ordering.
Table~\ref{tab:component_static_detail} isolates the same coordinates before
the two-score readout. First AC improves over DC by 20.8 accuracy points and
17.9 AUC points. Combining the separately computed DC and AC cosine scores
retains AC's accuracy and adds 0.4 AUC point. Thus the joint fingerprint is a
common measurement space. The task-specific readout determines which
coordinate matters. Global trajectory location dominates dynamic identity,
whereas coarse zero-mean evolution dominates static pair relationships.

\subsection{Matched Temporal-Representation Controls}
\label{app:matched_temporal_controls}

We isolate temporal encoding from proxy capacity with a single-pass
controlled experiment. At each selected proxy layer, one frozen forward pass
produces all compared representations from the identical effective response
tokens. DCT mode zero is exactly the all-token arithmetic mean. The same pass
also records the final-token state and coordinate-wise maximum. We evaluate
single-block controls of dimension $d$, pairwise concatenations of dimension
$2d$, and DCT bandwidths of two, four, and eight modes. Two learned controls
fit one or two global linear mixtures of the first eight modes using
the outer-training prompts. Let $\mathbf{z}_0,\ldots,\mathbf{z}_7$ denote
the standardized mode blocks. Head $h$ learns
$\mathbf{a}_h\in\mathbb{R}^{8}$ and outputs
\begin{equation}
\widetilde{\mathbf{z}}_h
=
\sum_{m=0}^{7}
\frac{a_{h,m}}{\lVert\mathbf{a}_h\rVert_2}\mathbf{z}_m.
\label{eq:learned_temporal_mixture}
\end{equation}
The one-head control returns one $d$-dimensional block. The two-head control
concatenates two such blocks to match the $2d$-dimensional DC--AC
representation. Each fold jointly fits these global mode weights and the
source probe. The weights are global and shared across responses, defining a
fixed temporal mixture. Every representation then uses the same
fold-local standardization, full-batch Adam linear probe, prompt-grouped
splits, and evidence accumulator.

\begin{table*}[t]
\centering
\small
\setlength{\tabcolsep}{5pt}
\begin{tabular*}{\textwidth}{@{\extracolsep{\fill}}lccrrrr@{}}
\toprule
\textbf{Temporal representation} & \textbf{Dim.} & \textbf{Learned?} &
\textbf{Acc.@1} & \textbf{M-F1@1} &
\textbf{Acc.@100} & \textbf{M-F1@100} \\
\midrule
All-token mean (DC)       & $d$  & No  & .475 & .470 & .956 & .955 \\
Final token               & $d$  & No  & .159 & .157 & .657 & .646 \\
Coordinate-wise maximum   & $d$  & No  & .410 & .407 & .928 & .926 \\
Mean + final              & $2d$ & No  & .386 & .387 & .896 & .894 \\
Mean + maximum            & $2d$ & No  & .478 & .476 & .947 & .945 \\
Final + maximum           & $2d$ & No  & .355 & .355 & .885 & .881 \\
Learned one-head mixture  & $d$  & Yes & .475 & .470 & .954 & .953 \\
Learned two-head mixture  & $2d$ & Yes & .478 & .476 & .953 & .952 \\
\midrule
DC--AC (DCT $q=2$)        & $2d$ & No  & .476 & .474 & .955 & .954 \\
DCT $q=4$                 & $4d$ & No  & .448 & .448 & .947 & .945 \\
DCT $q=8$                 & $8d$ & No  & .406 & .406 & .919 & .916 \\
\bottomrule
\end{tabular*}
\caption{\textbf{Strictly matched temporal representations on Agent500.}
Values average the four main proxies and three query-grouping seeds. All
features use the same proxy states, response-token span, fold-local selected
layer, and linear-probe protocol. ``Learned'' denotes additional fold-local
temporal mixture weights beyond the source probe.}
\label{tab:matched_temporal_controls}
\end{table*}
% Source:
% outputs/exp_agent_n500/100-way/matched_pooling_dct_modes_v1/

The matched controls localize the dynamic signal. DC--AC and
the exact mean differ by $+0.12$ point at $K=1$ (95\% CI
$[-0.10,+0.33]$) and $-0.05$ point at $K=100$ (95\% CI
$[-0.90,+0.80]$). Both confidence intervals include zero. The dimension-matched
mean--maximum concatenation and learned two-head mixture are likewise within
one point at both endpoints. The learned mixture is 0.18 point higher at
$K=1$ and 0.20 point lower at $K=100$, a negligible exchange despite the
extra fitted parameters.

Attribution declines as spectral bandwidth increases. Relative to
$q=2$, $q=4$ loses 2.76 points at $K=1$, while $q=8$ loses 6.99 points at
$K=1$ and 3.65 points at $K=100$. The $q=8$ losses remain significant after
Holm correction. Final-token, maximum, and the remaining fixed
concatenations are also consistently weaker. Across these controls, the frozen
proxy and its response-wide DC coordinate carry most dynamic source evidence.
The first AC supplies a fixed coarse trajectory coordinate at negligible
complexity and provides the strongest static pair signal
(Table~\ref{tab:component_static_detail}). Higher-frequency expansion adds
dimension while reducing accuracy.

For the endpoint tests above, we cluster predictions by outer fold and
source, draw 20,000 cluster-bootstrap replicates, and use 50,000 paired
cluster sign flips. Holm correction is applied across the listed
representation comparisons at $K\in\{1,100\}$.

\subsection{Prompt Context at the Proxy Input}

\begin{figure*}[t]
  \centering
  \includegraphics[width=\textwidth]{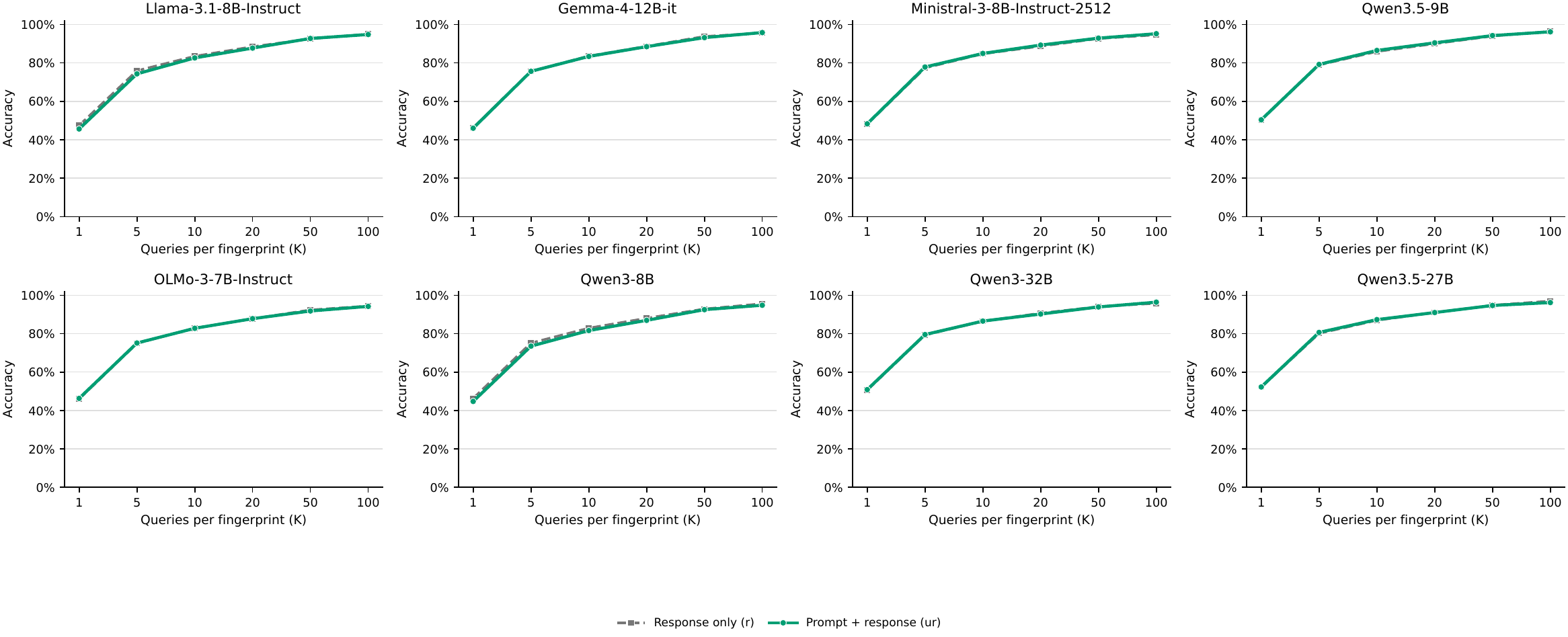}
  \caption{\textbf{Response-only and prompt--response proxy inputs.} Both
  variants encode response-token states. Their proxy forward passes differ in
  the inclusion of the observed prompt.}
  \label{fig:input_accuracy_full}
\end{figure*}

Figure~\ref{fig:input_accuracy_full} compares response-only input ($r$) with
prompt--response input ($ur$). Both apply the DCT to response-aligned states.
The curves nearly coincide across readers and budgets.
Response-only input averages 0.5 point higher at $K=1$, whereas
prompt--response averages 0.1 point higher at $K=100$. We report $ur$ in the
main tables because it retains the complete observed interaction and allows
the proxy to interpret a response in its query context. The near tie shows
that response content is sufficient for nearly all measured attribution
performance.

\subsection{Source-to-Prompt Variance}

For a held-out fingerprint coordinate, we use the balanced crossed-effects
decomposition
\begin{equation}
\mathbf{u}_{c,p}
=
\boldsymbol{\mu}
+\boldsymbol{\alpha}_{c}
+\boldsymbol{\beta}_{p}
+\boldsymbol{\epsilon}_{c,p},
\label{eq:source_prompt_decomposition}
\end{equation}
where $\boldsymbol{\alpha}_c$ and $\boldsymbol{\beta}_p$ are source and prompt
main effects. With one observation per source--prompt cell, the residual also
contains their interaction. We report
$R_{\mathrm{S/P}}=V_{\mathrm{source}}/V_{\mathrm{prompt}}$, averaged over
coordinates and outer folds.

\begin{figure*}[t]
  \centering
  \includegraphics[width=\textwidth]{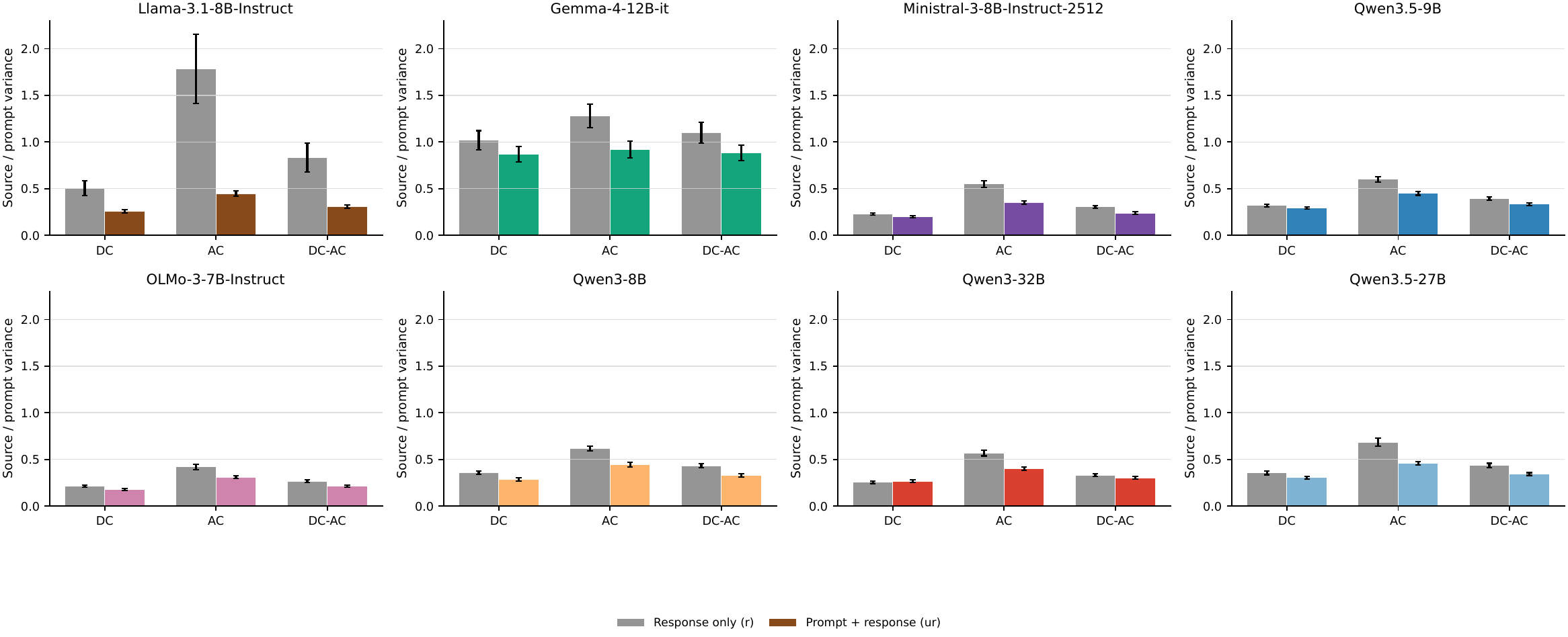}
  \caption{\textbf{Source-to-prompt variance across all proxy readers.}
  Bars show five-fold means and standard deviations for response-only and
  prompt--response inputs. The decomposition uses the held-out prompts.}
  \label{fig:source_prompt_input_full}
\end{figure*}

Figure~\ref{fig:source_prompt_input_full} shows that response-only input raises
$R_{\mathrm{S/P}}$ for every proxy and spectral block. First AC also has a
higher ratio than DC for most readers, a pattern consistent with its zero-sum
temporal weights attenuating response-wide offsets. The variance ratio and
attribution accuracy nevertheless diverge. Response-only and $ur$ are nearly
tied, whereas first AC trails DC on Agent500. The ratio diagnoses which
crossed factor dominates coordinate variance and should be interpreted
alongside held-out classification.

\subsection{Development-Stage Aggregation Alternatives}

During encoder development, we tested fixed-position pooling and supervised
position weights on the four main proxies under a uniformly sampled trajectory
protocol.
Table~\ref{tab:aggregation_design_pilot} separates three questions. First,
increasing uniform coverage from 16 to 64 positions produced the largest and
most consistent gain, peaking at $K=1$. Second, sampling every
fourth token from the response end was nearly tied with 32 uniformly spaced
positions, showing little sensitivity to the sampling grid. Third, the
Fisher-weighted variant assigned each of 16 relative positions a fold-local
one-way ANOVA score, normalized those scores into a single weight vector, and
reused that vector for every response. This is supervised static position
weighting with a global response-independent profile. Pure Fisher weighting
degraded every reported budget. Mixing 25\% Fisher weights with 75\% uniform
weights reduced both the loss and the magnitude of the effect.

Broader trajectory coverage supplied the most consistent gain in these
sampled-position controls. This result motivated the final encoder to use
every valid response token and retain fixed orthogonal DC--AC coordinates.

\begin{table}[t]
\centering
\footnotesize
\setlength{\tabcolsep}{1.5pt}
\begin{tabular*}{\columnwidth}{@{\extracolsep{\fill}}lrrr@{}}
\toprule
\textbf{Contrast} & \multicolumn{3}{c}{\textbf{Accuracy change (pp)}} \\
\cmidrule(l){2-4}
& \textbf{$K=1$} & \textbf{$K=50$} & \textbf{$K=100$} \\
\midrule
Uniform: M64 vs.\ M16            & +5.28 & +1.90 & +2.05 \\
Stride-4 vs.\ uniform M32        & $-$0.15 & +0.55 & +0.60 \\
Fisher vs.\ uniform (M16)        & $-$0.80 & $-$1.88 & $-$1.50 \\
25\% Fisher vs.\ uniform (M16)   & +0.49 & +0.10 & $-$0.15 \\
\bottomrule
\end{tabular*}
\caption{\textbf{Development-stage position-pooling diagnostics.} Positive
values favor the first configuration named in each row. Entries are
four-proxy mean accuracy changes in percentage points. Every row names its
matched baseline and defines an independent contrast.}
\label{tab:aggregation_design_pilot}
\end{table}

A separate first-fold projection pilot compared the full-rank two-mode
fingerprint with 64-dimensional PCA and class-centroid SVD variants. The
full-rank representation was best at $K=1$. Supervised SVD favored large
$K$ and used source labels to fit an additional projection. The final method
keeps the deterministic full-rank representation, avoids a second supervised
stage, and leaves source selection entirely to the linear probe. On the static
first-AC readout, PCA and GRP remain at or below the full-rank result
(Table~\ref{tab:component_static_detail}).

\FloatBarrier

\flushbottom

%% file: appendix/ecosystem_geometry.tex
\section{Model-Signature Geometry}
\label{app:model_geometry}

\subsection{Construction and Original-Space Enrichment}

For each proxy, geometry uses the modal layer among its five Agent500
outer-fold selections. A modal tie is resolved by mean inner-validation
accuracy across the five outer folds and then in favor of the shallower layer.
For each source $c$, we average the response-level fingerprints over the 500
Agent500 prompts. We then normalize the DC and first-AC blocks separately and
concatenate them, matching the geometry displayed in the main-paper figure.
Cosine distances are computed between the resulting 100 source centroids. The
primary family-purity analysis covers the seven provider families represented
by at least five models (87 models in total).
Purity aggregation uses these sufficiently populated groups, while neighbor
search retains all 100 models. Permutation tests randomly shuffle the seven
family assignments while preserving group sizes.

\begin{table}[t]
\centering
\footnotesize
\setlength{\tabcolsep}{2.5pt}
\begin{tabular*}{\columnwidth}{@{\extracolsep{\fill}}lrrrr@{}}
\toprule
\textbf{Proxy} & \textbf{Top-1} & \textbf{Top-3} &
\textbf{Top-5} & \textbf{Top-10} \\
\midrule
Llama-3.1-8B          & .816 & .632 & .494 & .360 \\
Gemma-4-12B           & .851 & .648 & .540 & .376 \\
Ministral-3-8B        & .828 & .670 & .559 & .401 \\
Qwen3.5-9B            & .851 & .693 & .563 & .418 \\
\midrule
OLMo-3-7B             & .851 & .693 & .586 & .424 \\
Qwen3-8B              & .839 & .690 & .549 & .391 \\
Qwen3-32B             & .839 & .697 & .600 & .414 \\
Qwen3.5-27B           & .874 & .682 & .579 & .426 \\
\bottomrule
\end{tabular*}
\caption{\textbf{Same-family neighbor purity in the original DC--AC cosine
space.} The random expectation is $18.4\%$. Values are proportions. The first
four rows are the main proxy roster.}
\label{tab:signature_family_purity}
\end{table}

All four main proxies exceed the permutation null at Top-1 and Top-5 after
Holm correction ($p\leq 8.0\times10^{-4}$ over eight tests). Their smallest
Top-5 enrichment is $2.69\times$. The six pairwise correlations between their
full $100\times100$ distance matrices range from $0.770$ to $0.933$, with
median Spearman correlation $0.818$. Their mean Top-10 neighborhood Jaccard
similarities range from $0.459$ to $0.618$. Family structure is therefore
present before dimensionality reduction and replicates across the four
proxies.

\subsection{Prompt and Projection Stability}

To test sensitivity to the response panel, we partition the 500 prompts into
five disjoint 100-prompt subsets and recompute every source centroid. Across
the four main proxies, Spearman correlations between subset and full-panel
distance matrices range from $0.940$ to $0.994$ (median $0.989$). The model
geometry is stable across disjoint prompt subsets.

For the two-dimensional visualization, we sweep five random seeds
($0$--$4$) and four perplexities ($10$, $15$, $30$, and $50$). Across these
20 runs, trustworthiness@10 ranges from $0.957$ to $0.973$ with median
$0.966$. Every run retains more than twice the random family purity at Top-5.
The displayed projection uses the separately fixed seed 42 and perplexity 15.
Interpretation is restricted to local neighborhoods because t-SNE axes and
global inter-cluster distances are arbitrary.

\subsection{Candidate Cross-Provider Neighborhoods}

We evaluate four candidate relationships first identified in the Qwen3.5-9B
projection using a cross-proxy rank analysis. A pair passes when
each model ranks the other within its Top-10 neighborhood for at least
three of four main proxies and the median mutual rank is at most 10.

\begin{table}[t]
\centering
\footnotesize
\setlength{\tabcolsep}{2.5pt}
\begin{tabular*}{\columnwidth}{@{\extracolsep{\fill}}p{0.64\columnwidth}rr@{}}
\toprule
\textbf{Model pair} & \textbf{Top-10} & \textbf{Med. rank} \\
\midrule
R1-Distill-Qwen-32B / QwQ-32B
  & 4/4 & 5.0 \\
Claude-Sonnet-4 / DeepSeek-V4-Pro
  & 4/4 & 2.5 \\
Gemma-4-31B / Gemini-3.1-Flash-Lite
  & 4/4 & 3.5 \\
MiniMax-M3 / Claude-Sonnet-4.6
  & 2/4 & 11.0 \\
\bottomrule
\end{tabular*}
\caption{\textbf{Cross-provider neighborhood stability across the four main
proxies.} Mutual rank is the worse of the two directed ranks.}
\label{tab:cross_provider_neighborhoods}
\end{table}

The distilled-Qwen relationship is stable under all eight proxies and aligns
with known base-model lineage. The other two passing relationships are
empirical cross-provider similarities whose lineage remains unestablished.
MiniMax-M3 and Claude-Sonnet-4.6 are immediate neighbors under Qwen3.5-9B and
several supplementary proxies, reaching the criterion for two of four main
proxies. The cross-proxy threshold separates stable neighborhoods from salient
single-proxy arrangements.

\subsection{Interactive Three-Dimensional Atlas}

\begin{figure}[H]
  \centering
  \includegraphics[width=\columnwidth]{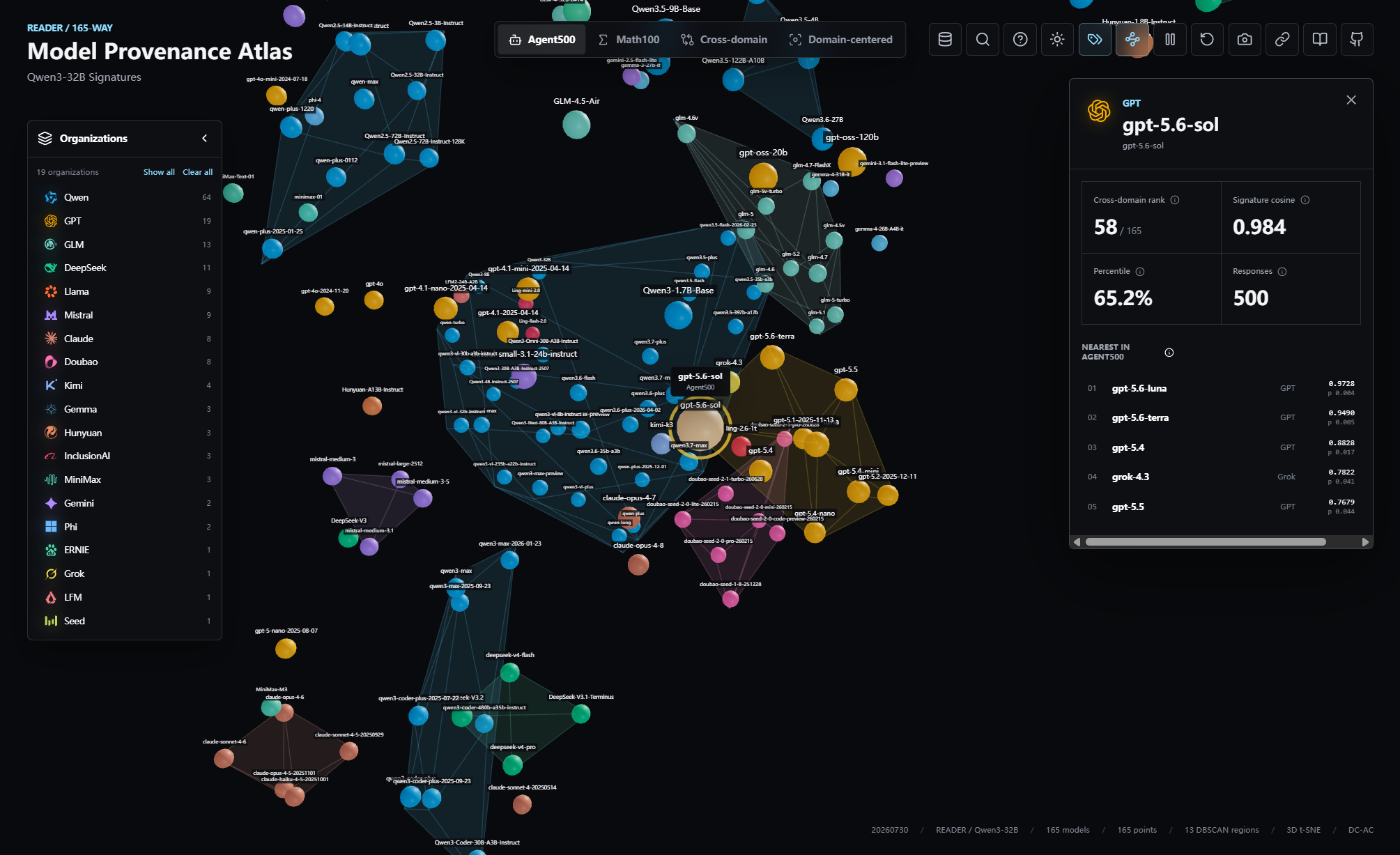}
  \caption{\textbf{Three-dimensional Model Provenance Atlas.} The screenshot
  shows Agent500 model signatures read by the Qwen3-32B READER proxy. The
  interface supports provenance-method, proxy, domain, organization, and model
  selection together with interactive neighborhood inspection.}
  \label{fig:model_provenance_atlas}
\end{figure}

Figure~\ref{fig:model_provenance_atlas} shows the visualization interface
developed to inspect model-signature geometry. It supports READER and LLM-DNA
representations, multiple proxy or encoder configurations, and Agent500,
Math100, cross-domain, and domain-centered views. Organization filters, model
search, and neighborhood inspection expose structure that is difficult to
read from a single static projection. The interface complements the paper
figures with interactive exploration. Quantitative neighborhood analyses use
the original signature space.

%% file: appendix/cross_domain_geometry.tex
\raggedbottom

\section{Cross-Domain Model-Signature Geometry}
\label{app:cross_domain_geometry}

\begin{figure*}[p]
  \centering
  \includegraphics[width=0.85\textwidth]{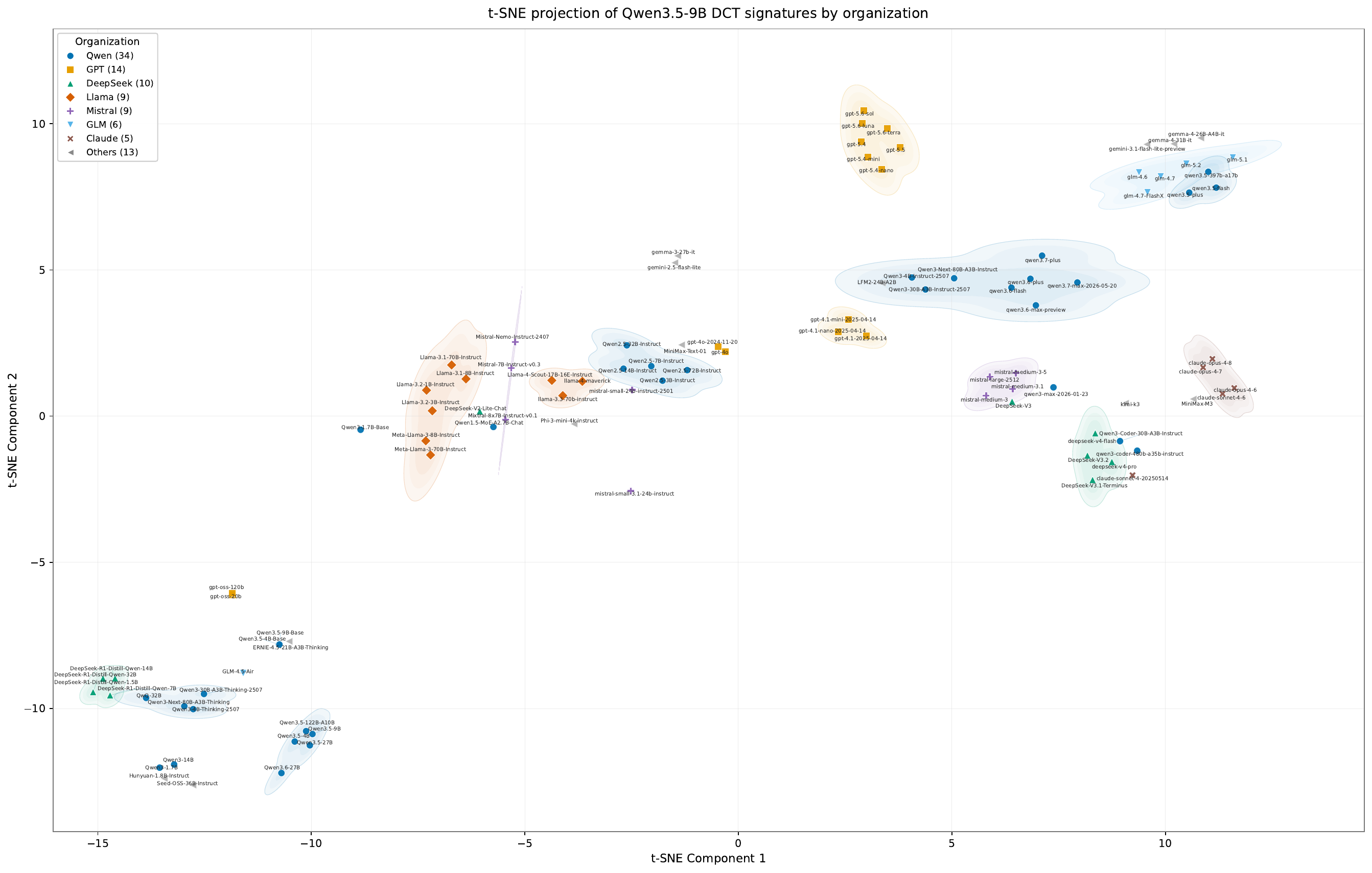}
  \caption{\textbf{Agent500 model-signature geometry.}
  The 100 endpoint centroids are computed from 500 heterogeneous interactions
  per endpoint. Contours visualize dense same-family regions. Quantitative
  results use the original fingerprint space.}
  \label{fig:model_signature_geometry_full}
  \vspace{1.5mm}
  \includegraphics[width=0.85\textwidth]{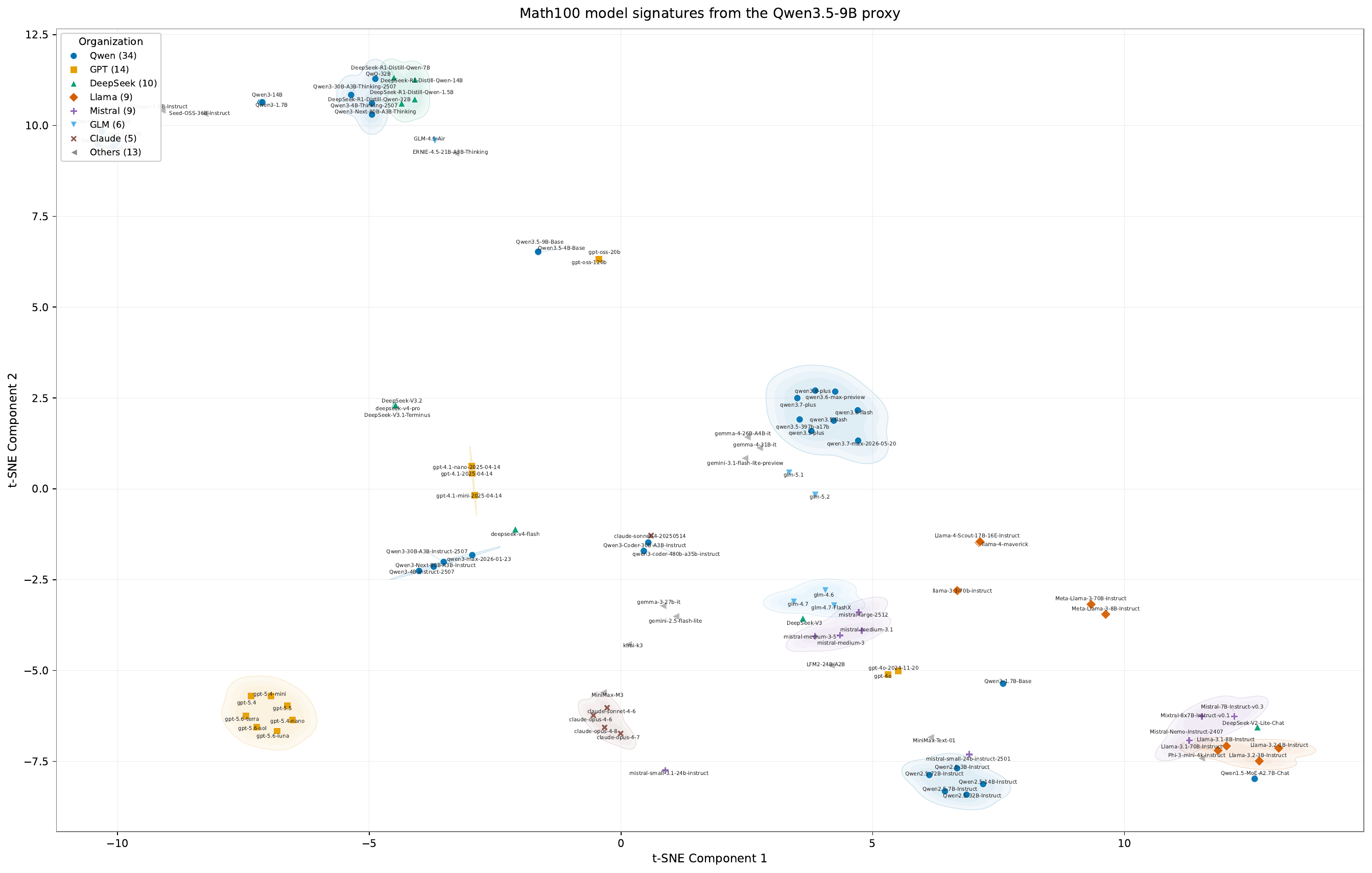}
  \caption{\textbf{Math100 model-signature geometry.}
  The same construction is applied to 100 mathematical interactions per
  endpoint. This projection is fit independently from
  Figure~\ref{fig:model_signature_geometry_full}.}
  \label{fig:model_signature_math100}
\end{figure*}

We examine model-signature geometry under a substantial shift in query domain.
The diagnostic uses the Qwen3.5-9B proxy at layer 18, its modal layer across
the five Agent500 outer-fold selections, together with the response-only input
view used for model-signature analysis. For each enrolled model endpoint, we
average its DC--AC fingerprints over 500 Agent500 responses or 100 Math100
responses, normalize the two coefficient blocks separately, and concatenate
them. Figures~\ref{fig:model_signature_geometry_full}
and~\ref{fig:model_signature_math100} fit t-SNE separately within each domain.
Independent fits assign unrelated axes and global cluster positions. Both use
cosine distance, seed 42, and perplexity 15.

\subsection{Geometry Within Each Domain}

The Math100 projection preserves substantial local organization despite the
narrower domain. Recent GPT endpoints, Claude endpoints, several Qwen
generations, and Llama endpoints form recognizable neighborhoods in both
panels. Their global arrangement changes, and several cross-family
neighborhoods are rearranged. Local family structure thus persists within
Math100. The joint and original-space analyses below quantify cross-domain
identity.

\subsection{Joint Geometry and Shared Domain Shift}

\begin{figure*}[p]
  \centering
  \includegraphics[width=0.85\textwidth]{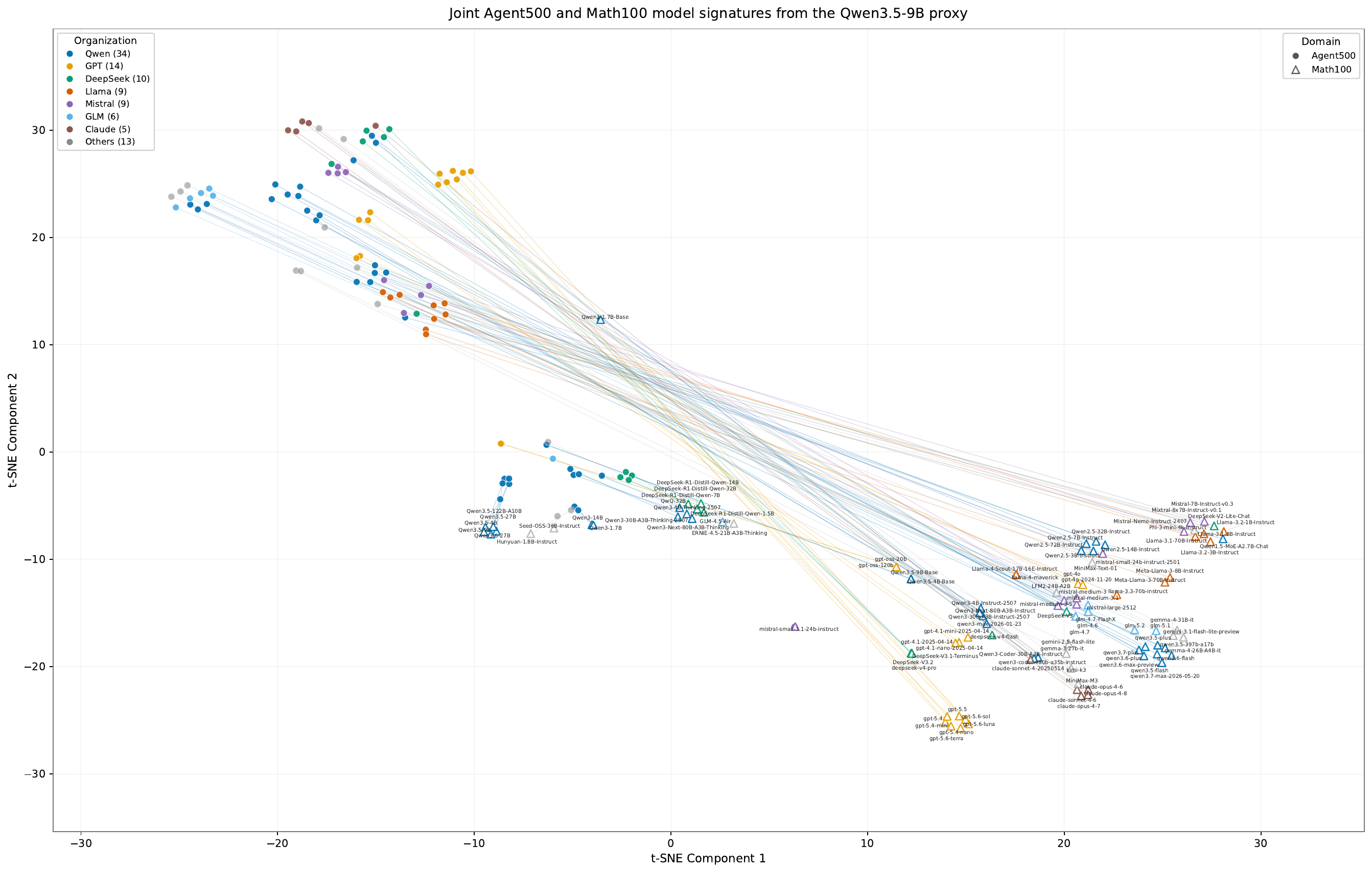}
  \caption{\textbf{Raw joint Agent500--Math100 projection.}
  Circles denote Agent500, triangles denote Math100, and lines connect the
  same model endpoint across domains. One t-SNE is fit to all 200 centroids.}
  \label{fig:model_signature_cross_domain_raw}
  \vspace{1.5mm}
  \includegraphics[width=0.85\textwidth]{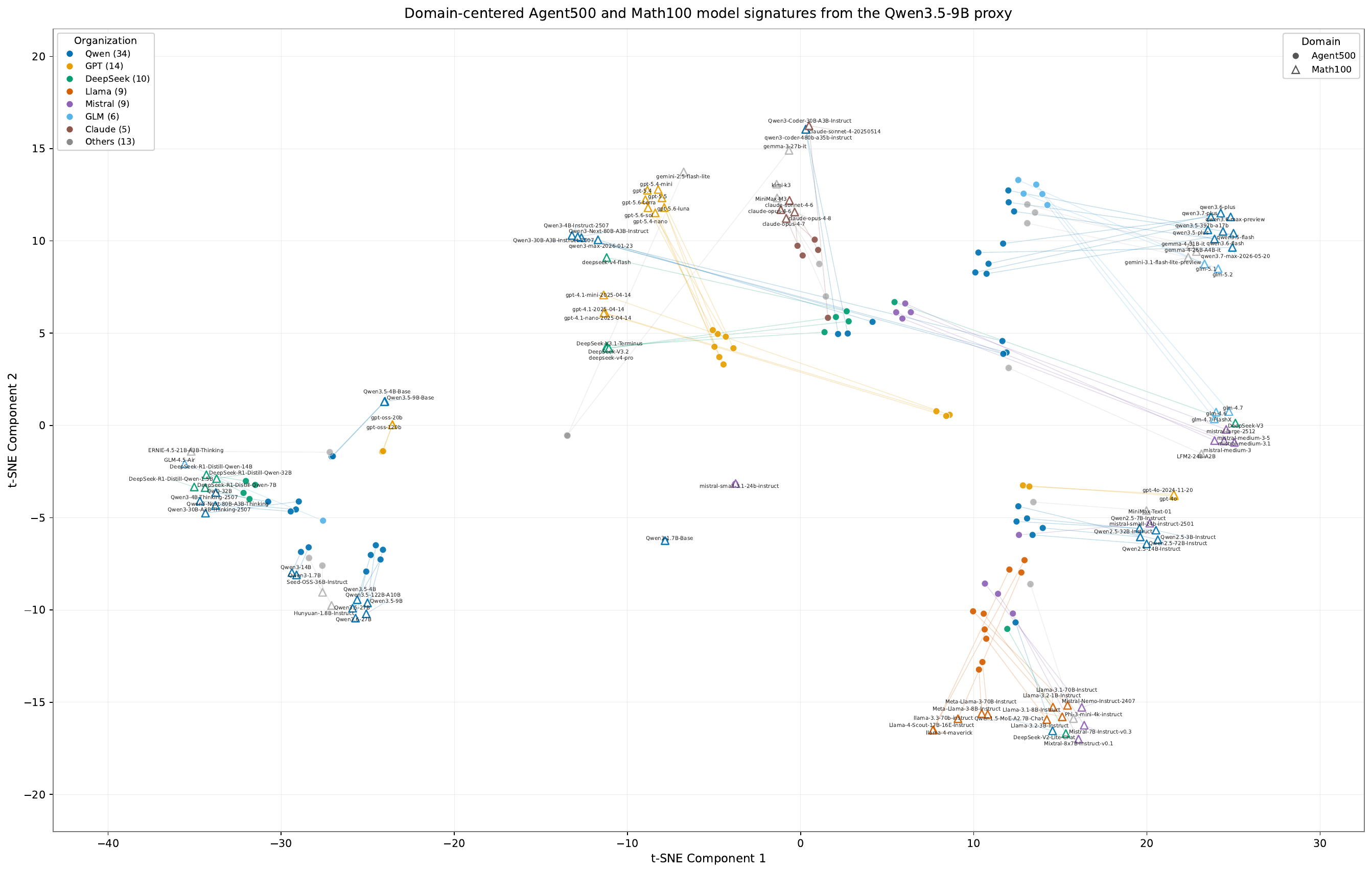}
  \caption{\textbf{Domain-centered joint projection.}
  Before fitting t-SNE, each domain mean is removed in the original
  fingerprint space and every endpoint signature is renormalized. Domain
  centering is used only for this geometry diagnostic.}
  \label{fig:model_signature_cross_domain_centered}
\end{figure*}

We next fit one t-SNE to all 200 centroids. Each line connects the Agent500 and
Math100 centroid of the same endpoint. In the raw joint projection
(Figure~\ref{fig:model_signature_cross_domain_raw}), circles and triangles
separate into two broad regions. The query domain thus contributes a shared
component large enough to dominate the two-dimensional neighborhood graph.
t-SNE distorts global metric distances, so line lengths serve as visual
guides.

To isolate residual structure, we perform one diagnostic transformation:
subtract the mean of the 100 endpoint centroids within each domain, then
renormalize every centered signature before the joint projection. This
unsupervised centering removes a domain-wide offset and is used only for this
geometry diagnostic. Figure~\ref{fig:model_signature_cross_domain_centered}
shows that many family-local neighborhoods align more closely after this
operation, whereas exact endpoint pairs remain dispersed.

\begin{table}[t]
\centering
\footnotesize
\setlength{\tabcolsep}{2.4pt}
\begin{tabular*}{\columnwidth}{@{\extracolsep{\fill}}lrrrrrr@{}}
\toprule
& \multicolumn{2}{c}{\textbf{Median cosine}} &
\multicolumn{3}{c}{\textbf{Same-model retrieval}} &
\textbf{Wins} \\
\cmidrule(lr){2-3}\cmidrule(lr){4-6}
\textbf{Space} & \textbf{Same} & \textbf{Other} &
\textbf{Top-1} & \textbf{Top-5} & \textbf{MRR} & \textbf{other} \\
\midrule
Raw       & .884 & .847  & .090 & .430 & .245 & .851 \\
Centered  & .416 & $-$.013 & .250 & .670 & .428 & .924 \\
\bottomrule
\end{tabular*}
\caption{\textbf{Cross-domain matching in the original fingerprint space.}
Each Agent500 centroid retrieves among all 100 Math100 centroids. ``Wins
other'' is the probability that the matched model has higher cosine
similarity than a randomly selected different model.}
\label{tab:cross_domain_signature_retrieval}
\end{table}

Table~\ref{tab:cross_domain_signature_retrieval} evaluates the original
8,192-dimensional fingerprints. Before centering, the matching
endpoint ranks first for 9 of 100 queries, yet it outranks 85.1\% of
different endpoints. Domain centering raises exact Top-1 matching to 25\%,
Top-5 matching to 67\%, and the pairwise win rate to 92.4\%. The cosine values
across the two rows have different baselines because centering changes the
coordinate origin.

Together, the projections and original-space retrieval metrics indicate a
two-level geometry. A broad query-domain component shifts the ecosystem as a
whole, while model- and family-specific residuals persist underneath it. This
pattern is consistent with Math100's within-domain organization and its lower
no-retraining attribution accuracy.

\FloatBarrier

\flushbottom

%% file: appendix/reproducibility.tex
\section{Reproducibility and Scope}
\label{app:reproducibility}

\subsection{Reproduction Checklist}

Table~\ref{tab:reproduction_map} maps each reported result to the information
needed to reproduce it. Released configurations record the proxy checkpoint,
layer-selection setting, input view, DCT modes, split seed, grouping seeds,
classifier settings, and evaluation protocol. Intermediate fingerprints are
keyed by benchmark, proxy, layer, and input view so that representation
extraction and statistical evaluation can be audited independently.

\begin{table}[H]
\centering
\footnotesize
\setlength{\tabcolsep}{3pt}
\begin{tabular*}{\columnwidth}{@{\extracolsep{\fill}}p{0.27\columnwidth}p{0.65\columnwidth}@{}}
\toprule
\textbf{Result family} & \textbf{Split, fitting, and repetitions} \\
\midrule
Agent500 &
Five outer prompt folds (seed 42), four-fold inner selection over all blocks by
response-level accuracy, fold-local standardizer/probe, and grouping seeds
42--44 at six budgets \\
Temporal controls &
Same Agent500 folds and Adam probe, one shared proxy pass, single- and
two-block pooling controls plus DCT $q\in\{2,4,8\}$ \\
Length shift &
Original prompt folds and frozen Agent500 estimators, with each response
assigned to the fold model that excluded its prompt \\
Math100 &
Five frozen fold models evaluated independently on 100 new prompts, with
metrics summarized by mean and standard deviation \\
Bench-A pairs &
Modal Agent500 layer, twenty fixed 92/24 pair splits, and a split-local
standardizer and SVM \\
Bench-A holdouts &
Twenty endpoint-disjoint splits and all 36 leave-two-family-out splits with
balanced train/test pairs and a split-local standardizer and SVM \\
Geometry &
Modal Agent500 layer, 500 prompts, original-space neighbor metrics, stored
prompt partitions, projection sweep, and permutations \\
\bottomrule
\end{tabular*}
\caption{\textbf{Reproduction map for the reported experiments.}}
\label{tab:reproduction_map}
\end{table}

\paragraph{Numerical determinism.}
The DCT coefficients and evidence grouping are deterministic once tokenizer
outputs and seeds are fixed. Floating-point proxy inference and GPU
optimization can introduce small implementation-level differences. We retain
per-fold response log-posteriors in addition to aggregate tables, allowing
all $K$-grouped metrics to be recomputed from cached scores.

\paragraph{API endpoints.}
Agent500 records the endpoint identifier and collection metadata for every API
response. The released responses and metadata define a fixed benchmark
snapshot as served endpoints evolve. Subsequent comparisons should preserve
the versioned identifiers in Table~\ref{tab:target_roster}.

\subsection{Computational Profile}

GPU-based representation extraction and runtime profiling were performed on
NVIDIA H200 accelerators with 140 GB HBM. Runtime measurements in
Table~\ref{tab:compute_profile} use a single GPU after warm-up.

Table~\ref{tab:compute_profile} separates learned capacity from the parameters
required by the audit forward path. READER fit time starts from cached
candidate-layer fingerprints and covers all inner-fold probe fits, layer
selection, and the final outer-training probe fit. Audit-time latency covers
proxy forward passes. Other frozen-representation methods start from cached
enrollment representations and include fold-local standardization. READER and
the DNA adaptations optimize their 100-way linear heads for 40 full-batch Adam
steps. LLMmap trains its $Q=1$ head
for six selected epochs. DeBERTa is fine-tuned end to end for two epochs.
Trainable parameter counts exclude standardizer statistics. Each inference
measurement covers tokenization, representation extraction, and the 100-way
score after model loading, with disk I/O excluded. We report the median
of three warmed runs over 256 interactions using each implementation's native
batch size. LLMmap's shared E5 encoder is counted once and covers both prompt
and response encoding. The dagger denotes adaptations from a
fixed-panel protocol to dynamic response-level attribution.

For READER, the inference count comprises the text embedding, Transformer
blocks through the selected hidden state, and the 100-way linear head.
The forward path exits at the selected block, before subsequent Transformer
blocks, the language-model output head, and unused vision modules. The latency
measurement uses the same early exit.

The frozen-proxy pass over response tokens dominates READER's cost and
parallelizes across proxy, source, and prompt. Once fingerprints are cached,
evidence regrouping and all query-budget evaluations reuse the same arrays.
Target models contribute text responses. The frozen proxy performs audit-time
feature extraction, and the source probe contains all trainable parameters.
Storage scales as $O(Nd)$ for two coefficients per response, and the probe
scales as $O(Cd)$. The full-rank design trades larger cached features for a
deterministic, projection-free representation.

The table reports Agent500 because its unit is an individual dynamic
interaction. MPT and PhyloLM operate on fixed-prompt model pairs in the static
Bench-A evaluation, whose pair-construction costs and per-example latency use
a different computational unit.

\subsection{Intended Scope}

READER targets dynamic auditing in an enrolled ecosystem through stateless
interactions. Its DC--AC fingerprint also supports static relationship
auditing within a shared measurement space. Dynamic identification uses the
source probe and evidence accumulator. Static model pairs use a relationship
readout over the same spectral blocks. The protocol assumes meaningful
candidate identities, labeled enrollment responses, a fixed ecosystem, and
independent requests. Unknown-source rejection, endpoint drift, conversational
state, and adversarial evasion remain open deployment conditions. The
no-retraining tests expose sensitivity to response coverage and domain shift.

Attribution may expose sensitive information. Deployments should document
collection policy, minimize stored text, restrict access to fingerprints and
source scores, and provide manual review for high-stakes decisions.